\documentclass[12pt]{article}
\usepackage{amsthm}
\usepackage{amssymb}
\usepackage{amsmath}
\usepackage{natbib}
\usepackage{graphicx}
\usepackage{float}
\usepackage[T1]{fontenc}
\usepackage{multirow}
\usepackage{appendix}
\usepackage[ruled,linesnumbered]{algorithm2e}
\usepackage{color}
\usepackage{rotating}
\usepackage{adjustbox}
\usepackage{diagbox}
\usepackage{makecell}
\usepackage{longtable}
\usepackage{subcaption}
\usepackage{url}
\usepackage{bm}
\usepackage{array}
\usepackage[colorlinks, linkcolor=blue, anchorcolor=blue, citecolor=blue]{hyperref}
\newcommand{\blind}{1}

\newtheorem*{remark*}{Remark}
\newtheorem*{lemma*}{Lemma}
\newtheorem{lemma}{Lemma}[section]
\newtheorem{theorem}{Theorem}[section]

\newtheorem{corollary}{Corollary}[section]

\newtheorem*{assumption*}{Assumption}
\newcommand{\aref}[1]{\hyperref[#1]{(#1)}}

\begin{document}

\def\spacingset#1{\renewcommand{\baselinestretch}%
	{#1}\small\normalsize} \spacingset{1}

\if1\blind
	{
		\title{Computational Efficient and Minimax Optimal Nonignorable Matrix Completion}
		\author{Yuanhong A \thanks{ayh@ruc.edu.cn} \\
			School of Statistics, Renmin University of China\\
			Guoyu Zhang \thanks{guoyz@stu.pku.edu.cn} \\
			Department of Probability and Statistics,
			School of Mathematical Sciences,  \\
			Center for Statistical Science, Peking University \\
			Yongcheng Zeng \thanks{zengyongcheng2022@ia.ac.cn}\\
			Institute of Automation, Chinese Academy of Sciences\\
			Bo Zhang \thanks{Corresponding author, mabzhang@ruc.edu.cn}\\
			School of Statistics, Renmin University of China\\
		}
		\date{}
		\maketitle
	} \fi

\if0\blind
	{
		\bigskip
		\bigskip
		\bigskip
		\begin{center}
			{\LARGE\bf Computational Efficient and Minimax Optimal Nonignorable Matrix Completion}
		\end{center}
		\medskip
	} \fi

\newcommand{\rank}{\operatorname{rank}}
\newcommand{\diag}{\operatorname{diag}}
\newcommand{\tr}{\operatorname{tr}}
\newcommand{\trace}{\operatorname{trace}}
\newcommand{\sign}{\operatorname{sign}}
\newcommand{\logit}{\operatorname{logit}}
\newcommand{\diam}{\operatorname{diam}}
\newcommand{\var}{\operatorname{var}}
\newcommand{\expit}{\operatorname{expit}}
\newcommand{\col}{\operatorname{col}}
\newcommand{\row}{\operatorname{row}}
\newcommand{\svd}{\operatorname{svd}}
\newcommand{\card}{\operatorname{card}}
\newcommand{\tor}{\operatorname{vector}}
\newcommand{\trun}{\operatorname{truncate}}
\newcommand{\mse}{\operatorname{MSE}}
\newcommand{\dis}{\operatorname{dist}}
\newcommand{\trexp}{\operatorname{trexp}}
\newcommand{\tol}{\operatorname{tol}}

\begin{abstract}%
	While the matrix completion problem has attracted considerable attention over the decades, few works address the nonignorable missing issue and all have their limitations. In this article, we propose a nuclear norm regularized row- and column-wise matrix U-statistic loss function for the generalized nonignorable missing mechanism, a flexible and generally applicable missing mechanism which contains both ignorable and nonignorable missing mechanism assumptions. The proposed method achieves computational efficiency comparable to the existing missing-at-random approaches, while providing the near minimax optimal statistical convergence rate guarantees for the more general nonignorable missing case. We propose an accelerated proximal gradient algorithm to solve the associated optimization problem, and characterize the interaction between algorithmic and statistical convergence. Simulations and real data analyzes further support the practical utility of the proposed method.
\end{abstract}

\noindent%
{\it Keywords:} missing not at random, matrix U-statistic, pairwise loss function, nuclear norm regularization, semiparametric generalized linear model
\vfill

\newpage
\spacingset{1.9} 

\section{Introduction}

Noisy matrix completion is a contemporary high-dimensional data problem that involves recovering a low-rank matrix from partial and noisy observations.  This challenging task has broad applications across various domains, including collaborative filtering \cite{srebro2004maximum, rennie2005fast}, computer vision \cite{zheng2012practical, zhou2014low}, and recommendation systems \cite{sindhwani2010one, ramlatchan2018survey}. For instance, in recommendation systems, the objective is to predict the unknown preferences of users for unobserved items based on a partially observed matrix.

The most common approach assumes the existence of a low-rank matrix parameter and estimates it by minimizing a loss function with matrix nuclear norm regularization, a method that has evolved over the years: \cite{ASVT10, MHT10, NOR11}. Since the matrix completion problem can be viewed as a missing data problem leveraging the low-rank characteristic of matrices as the core parameter, discussions on the missing mechanism are crucial. In early studies, most literature focused on the missing completely at random (MCAR) case \cite{ASVT10, MHT10, MCWN10, HLM11, NOR11}. Recently, the missing at random (MAR) mechanism has garnered significant attention \cite{WNT11, NW12, kloppNoisyLowrankMatrix2014a, IPS16, MCC18, MCL21, MCC23, MCL24}.

However, in the aforementioned literature, the missingness is assumed to be independent of the potential values themselves. In recommendation systems, these two components are usually related, where the missing mechanism is referred to as the nonignorable missing mechanism \cite{rubinInferenceMissingData1976}, which is also called the missing not at random (MNAR) mechanism. Taking movie rating data as an example, some individuals may already know they dislike certain movies through comments on the website or other methods, and thus, they choose not to watch these movies, naturally leaving the ratings blank. The imputation of these missing values allows us to recommend movies that users might prefer, but this is challenging because the observed samples themselves are biased, and the traditional methods mentioned above can only return results that deviate from the true recommendations \cite{little2019statistical}. The main project of this paper is to establish a unified framework to address this problem, thereby enabling a more reliable recommendation system.

The nonignorable missing mechanism has been established over decades in the context of regression problems; see \cite{NOI18} for an overview. However, extending these methods to the matrix completion problem is exceedingly challenging. \cite{ILM20}, \cite{MCC22}, and \cite{li2024a} have partially addressed this issue, but all approaches have their limitations. \cite{ILM20} considers a parametric missing mechanism and employs the expectation maximization (EM) method for estimation, which fails when the missing model is misspecified. Within the parametric framework, they do not resolve the model identification problem \cite{IDT14}, \cite{IDT23}, thus precluding the establishment of statistical guarantees. \cite{MCC22} adopts a semiparametric framework and provides statistical theory, but it necessitates the availability of an instrumental covariates tensor, which is unrealistic in many scenarios. \cite{li2024a} consider the similar nonignorable missing mechanism as we do, but they use the entire matrix U-statistic for estimation, leaving the \(O(n_1^2 n_2^2)\) computation complexity for each step of updating an \(n_1 \times n_2\) matrix, which is not feasible for high-dimensional matrix data. By leveraging the matrix structure, we propose the row- and column-wise matrix U-statistic type loss function, which has computation complexity \(O(n_1 n_2 \max\{n_1,n_2\})\), thus allowing us to handle the nonignorable missing mechanism for matrix completion without sacrificing computational efficiency.

In this paper, we address the matrix completion problem under the generalized nonignorable missing mechanism, a nonparametric unspecified nonignorable missing data mechanism assumption that is flexible and broadly applicable. Our contributions are threefold: (i)~we propose a computationally efficient low-rank matrix estimator for the generalized nonignorable missing mechanism in matrix completion problems, which maintains the same computational complexity as classical methods for MCAR / MAR case \cite{MHT10,NW12,MCL21}; (ii)~we resolve the identification problem for this missing mechanism through the proposed matrix U-statistic type loss function and a nuclear-norm-based matrix transformation; (iii)~we develop novel theoretical tools for analyzing matrix U-statistic, and establish the statistical property, show that our estimation error's Frobenius norm upper bound achieves a convergence rate comparable to existing MCAR / MAR missing mechanism results~\cite{koltchinskiiOracleInequalitiesEmpirical2011a,kloppNoisyLowrankMatrix2014a,MNC16}, matching the minimax optimal rate up to logarithmic factors.

The remainder of this paper is organized as follows: We first introduce the necessary notations in the remainder of this section. Section~\ref{sec:modelandmethod} presents the flexible generalized nonignorable missing mechanism and the corresponding estimation method. The theoretical guarantees of our estimator are established in Section~\ref{sec:theory}. Section~\ref{sec:numericalexperiments} demonstrates the empirical performance of our method through comprehensive simulation studies and real-world data analysis. Finally, we conclude with a discussion in Section~\ref{sec:conclusion}.

\textbf{Notation.} Given an $n_1 \times n_2 $ matrix $\bm{A} = (a_{ij})_{i,j = 1}^{n_1, n_2}$, we use $\|\bm{A}\|$, $\|\bm{A}\|_{\star}$, $\|\bm{A}\|_F$ to denote the spectral norm, the nuclear norm, and the Frobenius norm respectively. We also take  $\|\bm{A}\|_{\infty} $ is the vectorized infinity norm equal $\max_{i,j = 1}^{n_1,n_2} |a_{ij}| $. We take $\sigma_d(\bm{A}) $ as the $d $-th singular value of $\bm{A} $. Here we take $\bm{1}_{n}$ as the $n \times 1 $ vector with every entry equal $1 $. For a scalar $c \in \mathbb{R} $, we denote $\bm{A} \oplus c $ as $\bm{A} + c \bm{1}_{n_1} \bm{1}_{n_2}^\top $ and $\bm{A} \ominus c$ for $\bm{A} \oplus (-c)$.  We denote $a \wedge b = \min\{a,b\} $ and $a\vee b = \max\{a,b\} $. For non-asymptotic results, we use $C$ to denote a constant that may change from line to line.

\section{Generalized Nonignorable Missing Mechanism and Matrix Estimation}
\label{sec:modelandmethod}

In this section, we present the unified framework for matrix estimation under the generalized nonignorable missing mechanism, followed by the computation algorithm.

\subsection{The Observation Model}

We denote the $ n_1 \times n_2 $ partially observed matrix as $ \bm{X} = (X_{ij}) $, with the corresponding missing indicator matrix $ \bm{W} = (W_{ij}) $, where $ W_{ij} = 1 $ indicates that $ X_{ij} $ is observed, and $ W_{ij} = 0 $ indicates that $ X_{ij} $ is missing. We consider a generalized nonignorable missing mechanism, modeled through the conditional probability:
\begin{equation}
	\label{eq:missingmechanism}
	\mathbb{P}(W_{ij} = 1 \mid X_{ij}) = a_{ij} \pi(X_{ij}),
\end{equation}
where $\{a_{ij}\}$ are fixed but arbitrary constants and $\pi(\cdot)$ is an unknown common function. We term this formulation the generalized nonignorable missing mechanism as it incorporates both the nonignorable function $\pi(\cdot)$ and heterogeneous parameters $\{a_{ij}\}$. This provides greater flexibility than classical nonignorable missing models that assume a unique probability formulation $\mathbb{P}(W_{ij} = 1 \mid X_{ij})$ \cite{NOI18}.

A key feature of our approach is that we neither specify a parametric form for $\pi(\cdot)$ nor impose structural assumptions on $\{a_{ij}\}$, making the missing mechanism highly flexible. This generality allows our framework to accommodate various practical scenarios where the missingness pattern may depend on the underlying values in complex ways.

For matrix $\bm{X}$, we posit the existence of a low-rank parameter matrix $\bm{M} = (m_{ij})$ such that each entry $X_{ij}$ follows a semiparametric generalized linear model (GLM) with density:
\begin{equation}
    \label{eq:model}
    \mathbb{P}(x \mid m_{ij}, f) = \exp\left(x m_{ij} - b(m_{ij}, f) + f(x)\right),
\end{equation}
where $b(m_{ij}, f) = \log \left(\int \exp(x m_{ij} + f(x)) \, \mathrm{d} x\right)$ is the log-normalization function, $f(\cdot)$ represents an unspecified nuisance function, and $m_{ij}$ serves as both the matrix parameter of interest and the natural parameter in the exponential family distribution of $X_{ij}$. In contrast to classical GLMs \cite{mccullagh2019generalized} with prespecified $f(\cdot)$ (e.g., Gaussian with $f(x) = -x^2/2$, Bernoulli with $f(x) = 0$, or Poisson with $f(x) = -\log(x!)$) -- as required by existing generalized factor models \cite{MLF22,GFI23,MCL24} for handling multi-type data and capturing nonlinear $\bm{X}$-$\bm{M}$ relationship -- our framework permits a nonparametric structure for $f(\cdot)$, thus providing additional model flexibility.

From the density function \eqref{eq:model}, the likelihood ratio ${\mathbb{P}(x \mid m_1,f)}/{\mathbb{P}(x \mid m_2,f)}$ is monotonically increasing in $x$ when $m_1 > m_2$. This monotone likelihood ratio property implies that the distribution of $X_{ij}$ shifts to the right as $m_{ij}$ increases, making larger values more probable. Therefore, we can construct a recommendation system that prioritizes items with higher estimated $m_{ij}$ values, as these correspond to greater predicted user preference.

\subsection{The Proposed Estimator}
\label{sec:estimator}

Now we propose the estimator of \( \bm{M} \) under the nonignorable missing mechanism mentioned above. To ensure the low-rank structure, we still estimate $\bm{M}$ by optimizing the loss function with nuclear norm penalty regularization and matrix infinity norm constraint \cite{MHT10,kloppNoisyLowrankMatrix2014a,hamidiLowrankTraceRegression}:
\begin{equation}
	\label{eq:nuclearpenalty}
	\hat{\bm{M}} = \underset{\bm{M} \in \mathbb{R}^{n_1 \times n_2}, \|\bm{M}\|_{\infty} \leq \alpha} {\arg\min} \mathcal{L}(\bm{M}) + \lambda \|\bm{M}\|_\star,
\end{equation}
where $\lambda$ is a tuning parameter to be selected and $\mathcal{L}(\bm{M})$ denotes the loss function. To address both the complex generalized nonignorable missing mechanism in \eqref{eq:missingmechanism} and the semiparametric structural's nuisance components $b(\cdot, f)$ and $f(\cdot)$ in \eqref{eq:model}, we utilize the following pairwise pseudo-log-likelihood function, which is similar to the loss function in \cite{NPE13,LRT17,PPNM18} for regression framework:
\begin{equation}
	\label{eq:pseudolikelihood}
	\begin{aligned}
		 & l_{i_1j_1,i_2j_2}(m_{i_1j_1}, m_{i_2 j_2}) = \log(1 + \exp(-(X_{i_1 j_1} - X_{i_2 j_2})(m_{i_1 j_1} - m_{i_2 j_2}))),
	\end{aligned}
\end{equation}
one can see Supplement Material Section S.2 for the derivation.

Notice above loss function is pairwise, so a direct approach is to construct the matrix loss function utilizing every pair of observed element $(i,j), (i',j')$, as done in \cite{li2024a}:
\[
	\mathcal{L}_e(\bm{M}) = \frac{1}{n_1 n_2} \sum_{i,j} \sum_{i',j'} W_{ij} W_{i'j'} l_{ij, i'j'}(m_{ij}, m_{i'j'}),
\]
that for $m = \sum_{i,j} W_{ij}$ observed entries, computing $\mathcal{L}_e(\bm{M})$ requires $m(m-1)/2$ pairwise summations, yielding $O(m^2)$ computational complexity for the gradient matrix $\nabla \mathcal{L}_e(\bm{M}) = \partial \mathcal{L}_e(\bm{M})/\partial m_{ij}$. When $m = O(n_1n_2)$, this becomes $O(n_1^2n_2^2)$, creating a fourth-order scaling problem that is computationally prohibitive for high-dimensional matrix completion.

For a low-rank matrix $\bm{M}$ with $\rank(\bm{M})\leq d$, we consider the decomposition $\bm{M}=\bm{U}\bm{V}^\top$, where $\bm{U} \in \mathbb{R}^{n_1\times d}$ and $\bm{V} \in \mathbb{R}^{n_2 \times d}$. This allows estimation of $\bm{M}$ through its factor matrices. Notice that the $i$-th row of $\bm{U}$ depends only on the $i$-th row of $\bm{M}$, and similarly for $\bm{V}$'s $j$-th row and $\bm{M}$'s $j$-th column, we propose a row- and column-wise matrix U-statistic:
\begin{align}
	\mathcal{L}(\bm{M}) & = \sum_{i_1,j_1=1}^{n_1,n_2} W_{i_1j_1} \left( \frac{\sum_{j_2=1}^{n_2} W_{i_1j_2} l_{i_1j_1,i_1j_2}(m_{i_1j_1},m_{i_1j_2})}{n_2} + \frac{\sum_{i_2=1}^{n_1} W_{i_2j_1} l_{i_1j_1,i_2j_1}(m_{i_1j_1},m_{i_2j_1})}{n_1} \right) \nonumber \\
	& = \frac{1}{n_2} \sum_{i = 1}^{n_1}\sum_{j_1,j_2 = 1}^{n_2} W_{ij_1} W_{ij_2} l_{ij_1,ij_2}(m_{ij_1},m_{ij_2}) + \frac{1}{n_1} \sum_{j = 1}^{n_2} \sum_{i_1,i_2 = 1}^{n_1} W_{i_1j} W_{i_2j} l_{i_1j,i_2j}(m_{i_1j},m_{i_2j}), \nonumber \label{eq:matrixloss}
\end{align}
where for each observed entry $(i,j)$ (with $W_{ij} = 1$), estimation uses only data from the corresponding $i$-th row (first term) and $j$-th column (second term). This formulation reduces the computational complexity to $O(n_1n_2\max\{n_1,n_2\})$ for $\nabla \mathcal{L}(\bm{M})$, matching that of matrix SVD. Since the classical nuclear-norm penalty regularization problem \eqref{eq:nuclearpenalty}'s algorithm requires the gradient matrix $\nabla \mathcal{L}(\bm{M})$ and matrix SVD for updates, our approach maintains comparable computational efficiency.

By employing the pairwise pseudo-log-likelihood function \eqref{eq:pseudolikelihood}, our loss function $\mathcal{L}(\bm{M})$ remains invariant to the nuisance functions $f(\cdot)$, $\pi(\cdot)$ and parameters $\{a_{ij}\}$ in data generation model \eqref{eq:model} and missing mechanism \eqref{eq:missingmechanism}. This property enables our method to resolve the identification problem for the parameter matrix $\bm{M}$, allowing us to establish the consistency of the estimator $\hat{\bm{M}}$ \eqref{eq:nuclearpenalty}, as demonstrated in the following Section~\ref{sec:theory}.

Now we provide the convexity of our loss function. As function $l_{i_1j_1,i_2j_2}(m_{i_1j_1}, m_{i_2j_2})$ is convex in $m_{i_1j_1} - m_{i_2j_2}$, which implies that $\mathcal{L}(\bm{M})$ inherits this convexity property. Specifically, for any matrices \(\bm{M}_1\) and \(\bm{M}_2\) satisfying $\|\bm{M}_1\|_\infty, \|\bm{M}_2\|_\infty \leq \alpha$, we have:
\begin{equation}
	\label{eq:convexlowerbound}
	\begin{aligned}
		 & \mathcal{L}(\bm{M}_1) - \mathcal{L}(\bm{M}_2) \geq \tr(\nabla \mathcal{L}(\bm{M}_2)^\top (\bm{M}_1 - \bm{M}_2)) + \mathcal{D}_s^2(\bm{M}_1 - \bm{M}_2),
	\end{aligned}
\end{equation}
where $\mathcal{D}_s(\cdot)$ is a sample semi-norm for matrices defined through:
\begin{equation}
	\label{eq:seminorm}
	\begin{gathered}
		\mathcal{D}_s^2(\bm{M}) = \frac{1}{n_2}\sum_{i=1}^{n_1}\sum_{j_1,j_2=1}^{n_2}|m_{ij_1} - m_{ij_2}|^2 \mathcal{W}_{ij_1,ij_2} + \frac{1}{n_1}\sum_{j=1}^{n_2}\sum_{i_1,i_2=1}^{n_1}|m_{i_1j} - m_{i_2j}|^2 \mathcal{W}_{i_1j,i_2j},\\ 
	\end{gathered}
\end{equation}
with weights $\mathcal{W}_{i_1j_1,i_2j_2} $ given by:
\[
		\mathcal{W}_{i_1j_1,i_2j_2} = \frac{W_{i_1j_1} W_{i_2j_2}(X_{i_1j_1} - X_{i_2j_2})^2}{4 + 2 \exp(2\alpha(X_{i_1j_1} - X_{i_2j_2})) + 2 \exp(2\alpha(X_{i_2j_2} - X_{i_1j_1}))},
\]
the semi-norm $\mathcal{D}_s^2(\bm{M})$ arises naturally from the matrix U-statistic structure of $\mathcal{L}(\bm{M})$. We term it a semi-norm because $\mathcal{D}_s^2(\bm{M}) = 0$ does not imply $\bm{M} = \bm{0}$, and consequently, \eqref{eq:convexlowerbound} cannot be interpreted as a strong convexity condition. While for each weight $\mathcal{W}_{i_1j_1,i_2j_2}$, it provides a lower bound for half of the second derivative of $W_{i_1j_1} W_{i_2j_2} l_{i_1j_1,i_2j_2}(m_{i_1j_1},m_{i_2j_2})$ with respect to $m_{i_1j_1} - m_{i_2j_2}$ under the constraint $\|\bm{M}\|_\infty \leq \alpha$.

Given that $\mathcal{L}(\bm{M}) $ is a convex function \eqref{eq:convexlowerbound}, the feasible set for problem \eqref{eq:nuclearpenalty} is a convex set and the nuclear norm penalty is convex, it follows that problem \eqref{eq:nuclearpenalty} is a convex optimization problem. Consequently, we can employ the proximal gradient algorithm to solve this problem effectively.

\subsection{Optimization Algorithm}

We first introduce the proximal operator $\mathcal{S}_{\lambda,\alpha}(\bm{A})$ for an $n_1 \times n_2$ matrix $\bm{A}$, :
\begin{equation}
	\label{eq:proximalgradient}
	\mathcal{S}_{\lambda,\alpha}(\bm{A} ) = \underset{\| \bm{X}  \|_{\infty} \leq \alpha}{\arg\min} \frac{1}{2} \|\bm{X}  - \bm{A} \|_F^2 + \lambda \|\bm{X} \|_\star.
\end{equation}

Then based on the proximal gradient method \cite{beckGradientbasedAlgorithmsApplications2009, ASVT10}, we propose the following algorithm:
\begin{algorithm}[h]
	\caption{Proximal Gradient Algorithm}
	\label{alg:proximal}
	\KwData{Missing indicator $\bm{W}$, observed data $\bm{X}$, parameters $\lambda $ and $\alpha $.}
	\KwIn{Choose step size $\mu > 0$ and tolerance $\tol > 0$, randomly initialize matrix $\bm{M}_0$, compute $\mathcal{F}_0 = \mathcal{L}(\bm{M}_0) + \lambda \|\bm{M}_0\|_\star$.}
	\Repeat{$|\mathcal{F}_k - \mathcal{F}_{k+1}| < \tol$}{
		Compute $\bm{Y}_k = \bm{M}_k -  \nabla \mathcal{L}(\bm{M}_k) / \mu$;\\
		Update $\bm{M}_{k+1} = \mathcal{S}_{\lambda/\mu,\alpha}(\bm{Y}_k)$;\\
		Compute $\mathcal{F}_{k+1} = \mathcal{L}(\bm{M}_{k+1}) + \lambda \| \bm{M}_{k+1} \|_\star$;}
	\KwResult{Estimator matrix $\bm{M}_{k+1}$.}
\end{algorithm}

One can also use the FISTA (a fast iterative shrinkage-thresholding algorithm) \cite{beck2009fast} to accelerate this algorithm, that we update $\bm{M}_k$ by:
\begin{equation}
	\label{eq:fista}
	\begin{gathered}
		\bm{Z}_k = \bm{M}_k + \frac{(t_{k-1} - 1)}{t_k}  (\bm{M}_k - \bm{M}_{k-1}), \quad \bm{M}_{k + 1} = \mathcal{S}_{\lambda/\mu,\alpha}(\bm{Z}_k - \frac{1}{\mu} \nabla \mathcal{L}(\bm{Z}_k)), \\
		\text{where }t_1 = 1, \quad t_{k+1} = \frac{1 + \sqrt{1 + 4 t_k^2}}{2}.
	\end{gathered}
\end{equation}

To solve the optimization problem \eqref{eq:proximalgradient}, we use the two-block ADMM (alternating direction method of multipliers) algorithm. For the sake of narration, we first introduce the operators $\mathcal{S}_\tau^\star(\bm{A})$ and $\mathcal{S}_\alpha^t(\bm{A})$, that for an $n_1 \times n_2 $ matrix $\bm{A} = (a_{ij})_{i,j = 1}^{n_1,n_2}$ with $\svd(\bm{A}) = \bm{O}_1 (\diag(\sigma_1, \cdots, \sigma_{n_1 \wedge n_2})) \bm{O}_2^\top$:
\begin{align*}
	\mathcal{S}_\tau^\star(\bm{A}) = \bm{O}_1 (\diag((\sigma_1 & \! - \! \tau)_+, (\sigma_2 \! - \! \tau)_+, \cdots, (\sigma_{n_1 \wedge n_2} \! - \! \tau)_+)) \bm{O}_2^\top ,\ (\mathcal{S}_\alpha^t(\bm{A}))_{ij} & = (\! - \!\alpha) \vee  a_{ij} \wedge \alpha,\nonumber
\end{align*}
where $(x)_+$ is $x \vee 0$. Then the ADMM algorithm to solve \eqref{eq:proximalgradient} is:

\begin{algorithm}[h]
	\caption{ADMM Algorithm}
	\label{alg:admm}
	\KwData{Matrix $\bm{A}$, parameters $\lambda$ and $\alpha$.}
	\KwIn{Choose step parameter $\beta > 0$ and tolerance $\tol > 0$, randomly initialize matrices $\bm{X}_{1,0}, \bm{H}_0$, and use $\bm{X}_{2,0} = \mathcal{S}_\alpha^t(\bm{X}_{1,0})$.}
	\Repeat{$\max\{\| \bm{X}_{1,k} - \bm{X}_{2,k} \|_F, \|\bm{X}_{1,k} - \bm{X}_{1,k-1}\|_F, \|\bm{X}_{2,k} - \bm{X}_{2,k-1}\|_F\} < \tol$}{
	$\bm{X}_{1,k+1} =\mathcal{S}_{\lambda/(1+\beta)}^\star ((\bm{A} + \beta \bm{X}_{2,k} + \bm{H}_k)/(1 + \beta))$;
	$\bm{X}_{2,k+1} = \mathcal{S}_\alpha^t(\bm{X}_{1,k+1} - \bm{H}_k / \beta)$;
	$\bm{H}_{k+1} =  \bm{H}_k - \beta (\bm{X}_{1,k+1} - \bm{X}_{2,k+1})$;
	}
	\KwResult{Estimator matrix $\bm{X}_{2,k}$.}
\end{algorithm}

Note that when $\|\mathcal{S}^\star_\lambda(\bm{A})\|_\infty \leq \alpha$, the equality $\mathcal{S}_{\lambda,\alpha}(\bm{A}) = \mathcal{S}^\star_\lambda(\bm{A})$ holds. Applying the standard ADMM convergence theory, we establish that the sequence $\{\bm{X}_2^k\}$ converges to $\mathcal{S}_{\lambda,\alpha}(\bm{A})$ in problem \eqref{eq:proximalgradient}.

Here, we define the notations \(L_x, L_w \) and \( L_f \) as:
\begin{equation}
	\label{eq:Lf}
	\begin{gathered}
		L_x = (\max_{i,j:W_{ij} = 1} X_{ij} - \min_{i,j:W_{ij} = 1} X_{ij})^{2}, \  L_w =\frac{\max_{j} \sum_{i} W_{ij}}{n_1} + \frac{\max_{i} \sum_{j} W_{ij}}{n_2},\ L_f = \frac{L_x L_w}{2},
	\end{gathered}
\end{equation}
where $L_f $ plays as the Lipschitz constant of $\nabla \mathcal{L}(\bm{M}) $, then we have:

\begin{theorem}
	\label{th:algpro}
	For the proximal gradient algorithm (Algorithm~\ref{alg:proximal}) with $\mu > L_f$:
	\begin{enumerate}
		\item The sequence $\{\mathcal{F}_k\}$ is monotonically decreasing with:
		      \[
			      \mathcal{F}_{k} - \mathcal{F}_{k+1} \geq \frac{1}{2} (\mu - L_f) \|\bm{M}_k - \bm{M}_{k+1}\|_F^2.
		      \]
		\item The accumulation point of $\{\bm{M}_k \}$ is the optimal solution of \eqref{eq:nuclearpenalty}, with:
		      \[
			      \mathcal{F}_k - \mathcal{F}(\hat{\bm{M}}) \leq \mathcal{R}_k^P = \frac{\mu \|\bm{M}_0 - \hat{\bm{M}}\|_F^2}{2k},
		      \]
		      where $\mathcal{F}(\hat{\bm{M}}) = \mathcal{L}(\hat{\bm{M}}) + \lambda \|\hat{\bm{M}}\|_\star$.
	\end{enumerate}

	For the FISTA variant (Eq.~\eqref{eq:fista}), the convergence rate improves to:
	\[
		\mathcal{F}_k - \mathcal{F}(\hat{\bm{M}}) \leq \mathcal{R}_k^F = \frac{2\mu \|\bm{M}_0 - \hat{\bm{M}}\|_F^2}{(k+1)^2}.
	\]
\end{theorem}
This theorem establishes the sublinear convergence rate for both Algorithm~\ref{alg:proximal} and its FISTA variant \eqref{eq:fista}. Let $\mathcal{R}_k$ denote the algorithm error at iteration $k$, that equals $\mathcal{R}_k^P$ for Algorithm~\ref{alg:proximal} and $\mathcal{R}_k = \mathcal{R}_k^F$ for the FISTA variant \eqref{eq:fista}. Then the following Theorem~\ref{th:sparserelax} demonstrates that the $k$-th iterate $\bm{M}_k$ possesses the same statistical property as the estimator $\hat{\bm{M}}$ when $\mathcal{R}_k = O((n_1 + n_2)\mu)$. Considering $\| \bm{M}_0 - \hat{\bm{M}} \|_F^2 = O(n_1n_2) $, this implies the number of iterations satisfies $k \geq C(n_1 \wedge n_2)$ for Algorithm \ref{alg:proximal} and $k \geq C(\sqrt{n_1 \wedge n_2})$ for FISTA variant \eqref{eq:fista}, the complete statement and proof appear in the next section.

\section{Statistical Guarantee}
\label{sec:theory}

Here we denote $\bm{M}_\star$ as the true underlying parameter matrix, and $\hat{\bm{M}}$ is the estimator from \eqref{eq:nuclearpenalty}, then we will show the property of $\hat{\bm{M}}$ in this section. We first introduce the restricted nearly strong convexity (RNSC) condition to underpin the statistical rate: for the sample semi-norm $\mathcal{D}_s(\cdot)$ over a constraint set $\mathcal{C}$, there exists $\kappa_s > 0$:
\begin{equation}
	\label{eq:rsc}
	\mathcal{D}_s^2(\bm{\Delta}) \geq \kappa_s \mathcal{D}^2(\bm{\Delta}) \quad \text{for all } \bm{\Delta} \in \mathcal{C},
\end{equation}
where $\mathcal{D}(\cdot)$ denotes the population semi-norm obtained by setting all weights $\mathcal{W}_{i_1j_1,i_2j_2} \equiv 1$ in $\mathcal{D}_s(\cdot)$ (see Eq.~\eqref{eq:seminorm}). Unlike the classical matrix completion problem's well-known restricted strong convexity (RSC) condition \cite{hamidiLowrankTraceRegression} that $\mathcal{D}(\bm{\Delta}) = \|\bm{\Delta}\|_F$, our $\mathcal{D}(\cdot)$ is only a semi-norm satisfying $\mathcal{D}(\bm{M} \oplus c) = \mathcal{D}(\bm{M})$. This implies that the strong convexity of $\mathcal{L}(\bm{M})$ holds only up to an additive constant on set $\mathcal{C} $ -- the nearly strong convexity we term, presenting additional theoretical challenges. For the complete observation case ($W_{ij} \equiv 1$), we may take parameter $\kappa_s = \min \mathcal{W}_{i_1j_1,i_2j_2}$, demonstrating that condition \eqref{eq:rsc} properly relates the sample semi-norm $\mathcal{D}_s(\cdot)$ to its population counterpart $\mathcal{D}(\cdot)$. However, with missing data, this condition becomes unreliable over the full matrix space, necessitating our restriction to the carefully constructed subset $\mathcal{C}$.

Before introducing set $\mathcal{C}$, we first define the row and column spaces: for $n_1 \times n_2$ matrix $\bm{\Theta}$, $\row_r(\bm{\Theta})$ and $\col_r(\bm{\Theta})$ denote the orthogonal matrices corresponding to the top-$r$ right and left singular vectors, respectively:
\begin{gather*}
	\svd(\bm{\Theta}) = \bm{O}_1 \diag(\sigma_1(\bm{\Theta}), \sigma_2(\bm{\Theta}), \cdots, \sigma_{n_1\wedge n_2}(\bm{\Theta})) \bm{O}_2^\top, \\
	\col_r(\bm{\Theta}) = \bm{O}_1 \times \begin{pmatrix}
		\bm{I}_r \\
		\boldsymbol{0}
	\end{pmatrix}, \quad \row_r(\bm{\Theta}) = \bm{O}_2 \times \begin{pmatrix}
		\bm{I}_r \\
		\boldsymbol{0}
	\end{pmatrix},
\end{gather*}
then the constraint set $\mathcal{C}_r$ for RNSC \eqref{eq:rsc} is defined as:
\begin{equation}
	\label{eq:setc}
	\begin{gathered}
		\mathcal{C}_r = \{\bm{\Theta}:\|\bm{\Theta}_2\|_{\star} \leq 4 \sum_{k=r}^{n_1\wedge n_2} \sigma_k(\bm{M}_\star) + 3 \|\bm{\Theta}_1\|_\star \},
	\end{gathered}
\end{equation}
where $\bm{\Theta}_1, \bm{\Theta}_2 $ are the matrices:
\begin{gather*}
		\mathcal{M}_m(\bm{\Theta}) = \frac{\bm{1}_{n_1}^\top \bm{\Theta} \bm{1}_{n_2}}{n_1 n_2}, \quad  \bm{U} = \col_r(\bm{M}_\star \oplus \mathcal{M}_m(\bm{\Theta})),\quad \bm{V} = \row_r(\bm{M}_\star \oplus \mathcal{M}_m(\bm{\Theta})), \\
		\bm{\Theta}_2 = (\bm{I}_{n_1} - \bm{U U}^\top) (\bm{\Theta} \ominus \mathcal{M}_m(\bm{\Theta})) (\bm{I}_{n_2} - \bm{V V}^\top), \quad \bm{\Theta}_1 = \bm{\Theta} \ominus \mathcal{M}_m(\bm{\Theta}) - \bm{\Theta}_2, 
\end{gather*}
that $\mathcal{M}_m(\bm{\Theta})$ denotes the mean value of $\bm{\Theta}$. The matrix $\bm{\Theta}_2$ represents the high-rank component of $\bm{\Theta} \ominus \mathcal{M}_m(\bm{\Theta}) $ orthogonal to both the row and column spaces of $\bm{M}_\star \oplus \mathcal{M}_m(\bm{\Theta}) $'s top-$r$ singular vectors, while $\bm{\Theta}_1$ constitutes the low-rank part lying within these subspaces. The constraint set $\mathcal{C}_r$ \eqref{eq:setc} enforces a near low-rank structure of $\bm{\Theta} \ominus \mathcal{M}_m(\bm{\Theta})$: the high-rank component's nuclear norm $\|\bm{\Theta}_2 \|_\star$ is bounded by the low-rank part $\| \bm{\Theta}_1\|_\star $ and $\bm{M}_\star $'s tail singular value summation. In the Supplement Material, we prove that selecting $\lambda \geq 2\|\nabla\mathcal{L}(\bm{M}_\star)\|$ guarantees the estimation error $\hat{\bm{M}} - \bm{M}_\star$ lies in $\mathcal{C}_r$ for any $r \in \mathbb{N}^+$. 


Due to selection bias, consistent estimation of the mean value is unattainable in nonignorable missing data problems when both the data generation model and missing probability mechanism are unspecified \cite{little2019statistical}. Specifically for our framework, the function $f(\cdot)$ in the data generation model \eqref{eq:model}, the numbers $a_{ij}$ and function $\pi(\cdot)$ in the missing probability mechanism \eqref{eq:missingmechanism} remain unspecified, render $\bm{M}$ and $\bm{M} \oplus c$ statistical undistinguishable. Consequently, we can only identify the parameter matrix $\bm{M}$ up to a constant shift under the generalized nonignorable missing mechanism. Luckily, the constant shift preserves the relative ranking of entries of $\bm{M} $ that is needed for the recommendation system. As we take the nuclear-norm for regularization, so here we introduce the following transformation to solve the identification issue:
\begin{equation}
    \label{eq:definitionoftransformation}
	\hat{c} = \underset{c \in \mathbb{R}}{\arg\min} \| \bm{M} \oplus c \|_\star, \quad \mathcal{T}(\bm{M}) = \bm{M} \oplus \hat{c},
\end{equation}
which selects the representation with minimal nuclear norm, and the convexity of the nuclear norm guarantees the transformation's uniqueness, making $\mathcal{T}(\bm{M})$ identifiable for the generalized nonignitable missing mechanism (See Supplement Material). With this transformation, we introduce the following assumptions:
\begin{assumption*}
	\label{as:1}
	\begin{itemize}
		\item[\textcolor{blue}{(a)}\label{a}] Matrix $\mathcal{T}(\bm{M}_\star)$ satisfies $\|\mathcal{T}(\bm{M}_\star)\|_\infty \leq \alpha$.
		\item[\textcolor{blue}{(b)}\label{b}] The sample satisfies the RNSC condition \eqref{eq:rsc} on set $\mathcal{C}_r$ \eqref{eq:setc}.
	\end{itemize}
\end{assumption*}

While Assumption \aref{a} ensures the parameter matrix transformation $\mathcal{T}(\bm{M}_\star)$ is in the feasible set of optimization problem \eqref{eq:nuclearpenalty}. Assumption \aref{b} similar to the widely adopted RSC condition in matrix completion \cite{NW12, negahbanEstimationLowrankMatrices2011, kloppNoisyLowrankMatrix2014a,fanGeneralizedHighdimensionalTrace2019, hamidiLowrankTraceRegression} to derive the estimation error upper bound:

\begin{theorem}
	\label{th:oracleinequality}
	Under Assumptions \aref{a} and \aref{b}, when the tuning parameter $\lambda \geq 2 \|\nabla \mathcal{L}(\bm{M}_\star)\|$, the estimator $\hat{\bm{M}}$ \eqref{eq:nuclearpenalty} satisfies:
	\begin{gather*}
		\|\hat{\bm{M}} - \bm{M}_\star \ominus \mathcal{M}_m(\hat{\bm{M}} - \bm{M}_\star) \|_F \leq \max \left\{ \frac{3 \sqrt{2r} \lambda}{2 \kappa_s}, \quad \frac{4}{3 \sqrt{2r}} \sum_{k=r}^{n_1 \wedge n_2} \sigma_k(\bm{M}_\star) \right\}.
	\end{gather*}
\end{theorem}
This theorem shows the estimation error's upper bound with RNSC condition \eqref{eq:rsc}, aligns with results in the MAR setting -- Theorem 1 of \cite{fanGeneralizedHighdimensionalTrace2019} and Theorem 3.1 of \cite{hamidiLowrankTraceRegression}. It illustrates this upper bound depends on three key quantities: the sufficiently large tuning parameter $\lambda $, the rank parameter $r$ and the RNSC condition parameter $\kappa_s$ in \eqref{eq:rsc}. Notice that when $\bm{M}_\star$ is exactly low-rank, the $\bm{M}_\star $'s tail singular value summation term vanishes by taking $r = \rank(\bm{M}_\star) + 1$, yielding the following corollary:
\begin{corollary}
	\label{co:oracleinequality}
	When the sample $(\bm{X}, \bm{W}) $ satisfies the RNSC condition \eqref{eq:rsc} on $\mathcal{C}_{\rank(\bm{M}_\star)+1}$ and the tuning parameter $\lambda \geq 2 \|\nabla \mathcal{L}(\bm{M}_\star)\|$, with Assumption \aref{a}, the estimator $\hat{\bm{M}}$ \eqref{eq:nuclearpenalty} has:
	\[
		\| \hat{\bm{M}} - \bm{M}_\star \ominus \mathcal{M}_m(\hat{\bm{M}} - \bm{M}_\star) \|_F \leq \frac{3 \sqrt{2(\rank(\bm{M}_\star)+1)} \lambda}{2\kappa_s}.
	\]
\end{corollary}

To establish the convergence rate of $\|\hat{\bm{M}} - \bm{M}_\star \ominus \mathcal{M}_m(\hat{\bm{M}} - \bm{M}_\star) \|_F$ from Corollary~\ref{co:oracleinequality}, we need to: (i) derive a non-asymptotic bound for $2\|\nabla\mathcal{L}(\bm{M}_\star)\|$, which yields the optimal choice of $\lambda$; and (ii) verify the RNSC condition in \eqref{eq:rsc} with appropriate $\kappa_s$ or consider relaxing this requirement. In the sequel, we will first show the non-asymptotic rate of $\|\nabla\mathcal{L}(\bm{M}_\star)\|$ and then establish the convergence rate without Assumption \aref{b}.

Before presenting our main theoretical results, we first introduce two parameters governing the observation probabilities:
\begin{equation*}
	\pi_L := \min_{i,j} \mathbb{P}(W_{ij} = 1), \quad
	\pi_U := \max_{i,j} \mathbb{P}(W_{ij} = 1),
\end{equation*}
which bound the minimum and maximum observation probabilities, respectively. By definition, the number of observations $m$ satisfies: $n_1 n_2 \pi_L = O_p(m), m = O_p(n_1 n_2 \pi_U) $, a relationship that will be used for our subsequent theoretical results' discussions.

Here we first give the bound of $\|\nabla\mathcal{L}(\bm{M}_\star)\|$, which requires the following assumptions:
\begin{assumption*}
	\label{as:2}
	\begin{itemize}
		\item[\textcolor{blue}{(c)}\label{c}] The data generating process follows: (i) the density $\mathbb{P}(X_{ij} \mid m_{\star,ij}, f)$ follows model~\eqref{eq:model}, (ii) the missing mechanism $\mathbb{P}(W_{ij} = 1|X_{ij})$ follows~\eqref{eq:missingmechanism}, and (iii) the pairs $\{X_{ij},W_{ij}\}$ are mutually independent.

		\item[\textcolor{blue}{(d)}\label{d}] Denote $\tilde{X}_{i_1j_1,i_2j_2} = X_{i_1j_1} - X_{i_2j_2} $, $\tilde{m}_{\star,i_1j_1,i_2j_2} = m_{\star,i_1j_1} - m_{\star,i_2j_2} $ and $\tilde{Z}_{i_1j_1, i_2j_2}$ as:
		      \[
			     \frac{\tilde{X}_{i_1j_1,i_2j_2}}{2 + \exp(\tilde{X}_{i_1j_1,i_2j_2} \tilde{m}_{\star,i_1j_1,i_2j_2}) + \exp(-\tilde{X}_{i_1j_1,i_2j_2} \tilde{m}_{\star,i_1j_1,i_2j_2})} ,
		      \]
		      then for any $1 \leq i_1,i_2 \leq n_1 $ and $1 \leq j_1,j_2 \leq n_2 $, we require there exists $\alpha_{\psi_2} > 0 $:
		      \[
			      \mathbb{E}[\exp(\frac{\tilde{Z}_{i_1j_1,i_2j_1}^2}{\alpha_{\psi_2}^2} ) | W_{i_1j_1} = W_{i_2j_1} = 1]\leq 2,\ \mathbb{E}[\exp(\frac{\tilde{Z}_{i_1j_1,i_1j_2}^2}{\alpha_{\psi_2}^2} ) | W_{i_1j_1} = W_{i_1j_2} = 1] \leq 2.
		      \]
	\end{itemize}
\end{assumption*}

The Assumption \aref{c} is our model assumption in Section \ref{sec:modelandmethod}. For Assumption \aref{d}, the variables $\tilde{Z}_{i_1j_1,i_2j_2} $ represent $\partial l_{i_1j_1,i_2j_2}(m_1, m_2) /\partial m_1$ on $(m_{\star,i_1j_1},m_{\star,i_2j_2})$, which constitute the elements of the gradient matrix $\nabla \mathcal{L}(\bm{M}_\star) $. Since the function $g_m(x) = x/[2 + \exp(xm) + \exp(-xm)]$ is bounded by $1/(2|m|)$ and $|x|/4$, Assumption \aref{d} holds when the $X_{ij}$ is conditionally sub-Gaussian given $W_{ij}=1$ -- a standard requirement (see Assumption 9 in \cite{kloppNoisyLowrankMatrix2014a}, Condition C1 in \cite{fanGeneralizedHighdimensionalTrace2019}, and Theorems 4.1-4.2 in \cite{hamidiLowrankTraceRegression}). Moreover, for any data $\{X_{ij},W_{ij}\}$ satisfying Assumption \aref{c}, we can perform observation truncation to meet Assumption \aref{c} and \aref{d}, making this assumption particularly mild. Details appear in Supplement Material Section S.3.

Then the non-asymptotic probability bound for $\|\nabla \mathcal{L}(\bm{M}_\star)\|$ is established as follows:
\begin{lemma}
	\label{le:sparsespectralnorm}
	With Assumptions \aref{c} and \aref{d}, when $\pi_U \geq 2 {\log(n_1 \vee n_2)}/{(n_1 \wedge n_2)}  $, there exists a universal constant $C_1$:
	\[
		\mathbb{P}(\|\nabla \mathcal{L}(\bm{M}_\star)\| \geq C_1 \pi_U^{3/2} \alpha_{\psi_2} \sqrt{n_1\vee n_2}t) \leq 4(n_1 + n_2)\exp(-t^2) + \frac{2}{n_1 \vee n_2}.
	\]
\end{lemma}

By taking $t = \sqrt{2 \log(n_1 \vee n_2)} $ above, with probability at least $1 - {10}/{(n_1 \vee n_2)}$ we get:
\begin{equation}
	\label{eq:spectralnormprobound}
	\|\nabla \mathcal{L}(\bm{M}_\star)\| \leq C_1 \pi_U^{3/2} \alpha_{\psi_2} \sqrt{n_1 \vee n_2 \log(n_1 \vee n_2)},
\end{equation}
which matches Lemma 5 of \cite{kloppNoisyLowrankMatrix2014a}, where a term related to the observation rate multiplies factor $\sqrt{(n_1 \vee n_2)\log(n_1 \vee n_2)}$. Since our matrix $\nabla \mathcal{L}(\bm{M}_\star)$ takes the form of a matrix U-statistic, the standard matrix Bernstein inequality technique (Theorem 6.1.1 of \cite{tropp2015introduction}) is no more applicable. Establishing this result requires additional technical developments, which we present in the proof in Supplement Material.

It's noteworthy that the sample size requirement $m \geq C(n_1 \vee n_2)\log(n_1 \vee n_2)$ is typically required for establishing convergence rates in classical MCAR/MAR settings (Theorem 1 of \cite{NW12} and Theorem 10 of \cite{kloppNoisyLowrankMatrix2014a}). This aligns with our requirement $\pi_U \geq 2\log(n_1 \vee n_2)/(n_1 \wedge n_2)$ in Lemma \ref{le:sparsespectralnorm}, and it represents the weakest condition for sufficient observations which cannot be further relaxed.

As noted following Corollary \ref{co:oracleinequality}, establishing the convergence rate additionally requires verifying the RNSC condition \eqref{eq:rsc}, which relies on the following assumption:

\begin{assumption*}
	\begin{itemize}
		\item[\textcolor{blue}{(e)}\label{e}] There exists $\alpha_w > 0$ that for all $ 1 \leq i_1,i_2 \leq n_1 $ and $ 1 \leq j_1,j_2 \leq n_2$:
		      \[
			      \mathbb{E}[\mathcal{W}_{i_1j_1,i_2j_1} | W_{i_1j_1} = W_{i_2j_1} = 1]\geq \alpha_w,\  \mathbb{E}[\mathcal{W}_{i_1j_1,i_1j_2} | W_{i_1j_1} = W_{i_1 j_2} = 1] \geq \alpha_w.
		      \]
	\end{itemize}
\end{assumption*}

Notice that $\mathcal{W}_{i_1j_1,i_2j_2} $ is a strictly positive random variable conditional on $W_{i_1j_1} = W_{i_2j_2} = 1 $, thus this assumption is very mild. As $\mathbb{E}[\mathcal{W}_{i_1j_1, i_2j_2}] = \mathbb{E}[\mathcal{W}_{i_1j_1,i_2j_2} | W_{i_1j_1} = W_{i_2j_2} = 1 ] \mathbb{E}[W_{i_1j_1}] \mathbb{E}[W_{i_2j_2}] $, the Assumption \aref{e} implies $\mathbb{E}[\mathcal{W}_{i_1j_1,i_2j_2}] \geq \pi_L^2 \alpha_w $. Similarly to the techniques in \cite{kloppNoisyLowrankMatrix2014a}'s Lemma 12, this result is useful to build the RNSC condition \eqref{eq:rsc} under a specific matrix set, as we show in the following lemma:
\begin{lemma}
\label{le:rscmain}
	We denote the constraint set $\mathcal{C}(d) \subset \mathbb{R}^{n_1 \times n_2}$ with a universal constant $C_2 > 0$:
	\[
		\mathcal{C}(d) := \{\bm{A}:\mathcal{M}_m(\bm{A}) = 0, \|\bm{A}\|_\star \leq \sqrt{d} \|\bm{A}\|_F, \| \bm{A} \|_F \geq \frac{9 \log^2(n)(n d \pi_U^3 C_2)^{1/2}}{\alpha^2 \pi_L^2 \alpha_w}  \| \bm{A}\|_\infty \},
	\]
	where $n = n_1 + n_2 $. Then under Assumption \aref{c}, \aref{e} and $\pi_U \geq 1 / (n_1 \wedge n_2) $, with probability at least $1 - 4/n $, the RNSC condition \eqref{eq:rsc} holds on $\mathcal{C}(d) $ with $\kappa_s = \pi_L^2 \alpha_w/2$.
\end{lemma}

Although Theorem~1 of \cite{NW12}, Lemma~12 of \cite{kloppNoisyLowrankMatrix2014a}, and Theorem~3.2 of \cite{hamidiLowrankTraceRegression} establish similar results, their analyses are restricted to independent sampling matrices and rely on Massart's inequality (Theorem~14.2 of \cite{buhlmannStatisticsHighdimensionalData2011}) for probability bounds. These techniques cannot be directly applied to our setting because: previous work assumes independent sampling with a fixed sample size $m$ (the observation number), which facilitates the use of Massart's inequality. In contrast, our framework considers the U-statistic $\mathcal{W}_{i_1j_1,i_2j_2}$ derived from $\{X_{ij}, W_{ij}\}$ pairs, and the sample size should be viewed as $n_1 n_2 $ instead of $m $. This setting introduces substantial challenges in deriving concentration inequalities, which we overcome by employing the decoupling technique \cite{delapenaDecoupling1999} and careful moment control. See the proof in Supplement Material for details.

With the above preparations, we establish the following convergence rate result:

\begin{theorem}
\label{th:sparserelax}
Under Assumptions \aref{a}, \aref{c}, and \aref{e}, when $\lambda \geq 2\|\nabla\mathcal{L}(\bm{M}_\star)\|$ and $\pi_U \geq (n_1 \wedge n_2)^{-1}$, there exists a universal constant $C_3$ such that for any $r \in \mathbb{N}^+$, an event $E$ occurs with probability $\mathbb{P}(E) \geq 1 - 4/(n_1 + n_2)$ satisfying:
\begin{equation}
\label{eq:probabilityupperbound}
\begin{gathered}
\|\hat{\bm{M}} - \bm{M}_\star \ominus \mathcal{M}_m(\hat{\bm{M}} - \bm{M}_\star)\|_F \leq \max\left\{ 
\frac{3\sqrt{2r}\lambda}{\pi_L^2\alpha_w}, \quad \frac{4}{3\sqrt{2r}}\sum_{k=r}^{n_1 \wedge n_2}\sigma_k(\bm{M}_\star),\right. \\
\quad \left. C_3\frac{\sqrt{(n_1+n_2)r}\pi_U^{3/2}}{\alpha\pi_L^2\alpha_w}\log^2(n_1+n_2)
\right\}, \text{ on } E.
\end{gathered}
\end{equation}

Furthermore, under Assumption \aref{d} with $\pi_U \geq 2\log(n_1 \vee n_2)/(n_1 \wedge n_2)$ and tuning parameter $\lambda \geq 2C_1\pi_U^{3/2}\alpha_{\psi_2}\sqrt{2(n_1 \vee n_2)\log(n_1 \vee n_2)}$, we obtain the minimax upper bound:
\begin{equation}
\label{eq:upperbound}
\mathbb{E}\left[\|\hat{\bm{M}} - \bm{M}_\star \ominus \mathcal{M}_m(\hat{\bm{M}} - \bm{M}_\star)\|_F\right] \leq \text{RHS of \eqref{eq:probabilityupperbound}} + 56\alpha.
\end{equation}

For the algorithmic sequence $\{\bm{M}_k\}$ with error $\{\mathcal{R}_k\}$ as in Theorem \ref{th:algpro}, with the same conditions for probability bound \eqref{eq:probabilityupperbound}, the same event $E$ yields:
\begin{equation}
\label{eq:algorithmerrorbound}
\|\bm{M}_k - \bm{M}_\star \ominus \mathcal{M}_m(\bm{M}_k - \bm{M}_\star)\|_F \leq \text{RHS of \eqref{eq:probabilityupperbound}} + \frac{2\mathcal{R}_k}{3\sqrt{2r}\lambda}, \ \forall k \in \mathbb{N}^+ \text{ on } E.
\end{equation}
\end{theorem}

This theorem provides both probabilistic and expectation bounds for the estimation error $\|\hat{\bm{M}} - \bm{M}_\star \ominus \mathcal{M}_m(\hat{\bm{M}} - \bm{M}_\star)\|_F$ without requiring the RNSC condition \eqref{eq:rsc}. The probability bound \eqref{eq:probabilityupperbound} comprises three components: (i) the first two terms originate from Theorem \ref{th:oracleinequality} with $\kappa_s = \pi_L^2\alpha_w/2$, and (ii) the third term controls matrix behavior on $\mathcal{C}^c(98r)$ via Lemma \ref{le:rscmain}. The expectation bound \eqref{eq:upperbound} follows from taking expectations in \eqref{eq:probabilityupperbound} and applying a union bound argument for the estimation error's Frobenius norm. The algorithmic bound \eqref{eq:algorithmerrorbound} shows that each iterate $\bm{M}_k$ inherits the estimation error bound from \eqref{eq:upperbound} plus an additional algorithmic error term $2\mathcal{R}_k/(3\sqrt{2r}\lambda)$.

Analogous to Corollary \ref{co:oracleinequality}, when $\rank(\bm{M}_\star) \leq d$, setting $r = d + 1$ in Theorem \ref{th:sparserelax} eliminates $\bm{M}_{\star} $'s tail singular value summation term. When $\pi_U/\pi_L = O(1)$; $d, \alpha, \alpha_{\psi_2}, \alpha_w$ treated as constants; $L_w = O_p(\pi_U)$, $L_x = O_p(1)$ and $\mu = L_w L_x = O_p(\pi_U)$ in \eqref{eq:Lf}; $\lambda = 2\|\nabla\mathcal{L}(\bm{M}_\star)\| = O_p(\pi_U^{3/2} \sqrt{n_1 \vee n_2 \log(n_1 \vee n_2)})$ for optimal value. The upper bound \eqref{eq:algorithmerrorbound} indicates that when $\mathcal{R}_k/\lambda$ matches the order of the third term in \eqref{eq:probabilityupperbound}, the $k$-th iterate $\bm{M}_k$ achieves the same statistical convergence rate as $\hat{\bm{M}}$, specifically requires $\mathcal{R}_k = O_p(\sqrt{n_1 \vee n_2} \mu) $ -- needs $C n_1 \wedge n_2$ iterations for Algorithm \ref{alg:proximal} and $C \sqrt{n_1 \wedge n_2} $ for its FISTA variant \eqref{eq:fista}.

For the minimax optimality analysis, as the observation number satisfies $m = O_p(n_1 n_2 \pi_L)$, classical MCAR/MAR estimators \cite{koltchinskiiOracleInequalitiesEmpirical2011a,kloppNoisyLowrankMatrix2014a,MNC16} achieve an expectation upper bound of $O(\sqrt{n_1 n_2 (n_1 \vee n_2) d \log(n_1 \vee n_2)/m})$. Our upper bound \eqref{eq:upperbound} matches this rate up to a $\log^{3/2}(n_1 + n_2)$ factor, accounting for the matrix U-statistic complexity and random observation count $m$ (fixed in prior work). Additionally, the minimax lower bound for MCAR case is established in Theorem 6 of \cite{koltchinskiiOracleInequalitiesEmpirical2011a} gives $O(\sqrt{n_1 n_2 (n_1 \vee n_2) d/m}) $, which remains valid for our framework. Compared to this lower bound, our estimation error's upper bound \eqref{eq:upperbound} differs by up to a $\log^2(n_1 \vee n_2)$ factor, achieving the near minimax optimal rate.

In denser observation regimes, the upper bound \eqref{eq:upperbound} admits a sharper characterization. Specifically, when $\pi_U^3 \geq \log(n_1 + n_2) / (n_1 \wedge n_2)$, the expectation upper bound is reduced to $O(\sqrt{(n_1 + n_2)d\log(n_1 + n_2)} \pi_U^{3/2}/ \pi_L^2) $, matching the classical MAR matrix completion rate. Furthermore, when $\pi_L \geq c > 0 $ for some constant $c $, we can eliminate the logarithmic factors appear above, yielding the exact minimax optimal rate. Complete statements appear in Section S.4 of the Supplement Material.

Notice that all above theories focus on the difference between $\hat{\bm{M}} \ominus \mathcal{M}_m(\hat{\bm{M}}) $ and $\bm{M}_\star \ominus \mathcal{M}_m(\bm{M}_\star) $, reflecting the non-identifiability of $\mathcal{M}_m(\bm{M}_\star) $. While the transformation \eqref{eq:definitionoftransformation} can solve the identification problem, so we can control $\mathcal{T}(\hat{\bm{M}}) - \mathcal{T}(\bm{M}_\star) $ directly: here we denote $d = \rank(\mathcal{T}(\bm{M}_\star)) $, $\bm{U} = \col_d (\mathcal{T}(\bm{M}_\star)) $, $\bm{V} = \row_d (\mathcal{T}(\bm{M}_\star)) $ and $ \mathfrak{B}(\bm{M}_\star)$ as:
\begin{equation}
	\label{eq:bcdefintion}
	\begin{gathered}
		\mathfrak{B}(\bm{M}_\star) = \frac{\| (\bm{I}_{n_1} - \bm{U U}^\top) \bm{1}_{n_1} \| \| (\bm{I}_{n_2} - \bm{V V}^\top) \bm{1}_{n_2} \|}{\sqrt{n_1 n_2}} - \sqrt{n_1 n_2} |\mathcal{M}_m(\bm{U V}^\top)|,
	\end{gathered}
\end{equation}
then through the property of nuclear norm, we have the following theorem:
\begin{theorem}
	\label{th:thransform}
	Under Assumption \aref{a}, when $\lambda \geq 2 \|\nabla \mathcal{L}(\bm{M}_\star)\|$, for any $r \in \mathbb{N}^+$, we have:
	\begin{gather*}
		\| \mathcal{T}(\hat{\bm{M}}) - \mathcal{T}(\bm{M}_\star) \|_F \leq
		 \frac{9 \sqrt{2r} \|\hat{\bm{M}}  -  \bm{M}_\star \ominus \mathcal{M}_m(\hat{\bm{M}}-\bm{M}_\star) \|_F + 8\sum_{k=r}^{n_1\wedge n_2} \sigma_k(\bm{M}_\star) }{\mathfrak{B}(\bm{M}_\star)}.
	\end{gather*}
	
	Additionally, when $\|\hat{\bm{M}}\|_\infty < \alpha$, we have $\mathcal{T}(\hat{\bm{M}}) = \hat{\bm{M}}$.
\end{theorem}
Building on our previous results, we can establish Frobenius norm bound $\|\mathcal{T}(\hat{\bm{M}}) - \mathcal{T}(\bm{M}_\star)\|_F$ directly. Following Corollary \ref{co:oracleinequality}, the $\bm{M}_\star $'s tail singular values summation part vanishes by setting $r = \rank(\bm{M}_\star) + 1$. As we can always take $\alpha$ large enough, so without loss of generality, we can always consider $\hat{\bm{M}} = \mathcal{T}(\hat{\bm{M}})$. Theorem \ref{th:thransform} thus shows that our estimator $\hat{\bm{M}} $ is a good approximation of the parameter matrix's transformation $\mathcal{T}(\bm{M}_\star)$.

\section{Numerical Experiments}
\label{sec:numericalexperiments}

In this section, we demonstrate the performance of our method across two simulation settings and two real data sets by evaluating the estimation accuracy and metrics for ranking estimation performance. For our method, we take $\alpha = 10 $ in \eqref{eq:nuclearpenalty} and $\mu = \{L_x L_w / 2, 1.1\}$ in Algorithm \ref{alg:proximal} by choosing the $95\% $ and $5\% $ quantiles instead of the maximum and minimum values in $L_x, L_w $'s definition \eqref{eq:Lf}. The tuning parameters for our and the other baseline methods are selected to optimize performance for the corresponding metrics.

\subsection{Simulations}

For given sample size $n $, we generate matrix $\bm{M} $ as:
\begin{gather*}
	\bm{M} = \frac{1}{\sqrt{3}}[\mathcal{N}(0,1)]_{n \times 3} [\mathcal{N}(0,1)]_{3 \times n},
\end{gather*}
where $[\mathcal{N}(0,1)]_{m \times n} $ is the $m \times n $ matrix with each entry is sampled from the standard normal distribution independently. Here we consider the following two data-generating processes (DGP) for the matrix completion problem:
\begin{list}{}{
		\setlength{\itemindent}{0pt}
		\setlength{\leftmargin}{0pt}
	}
	\item \textbf{DGP1:} $X_{ij} \sim \mathcal{B}(\expit(m_{ij})) $, where $\mathcal{B}(p) $ is the Bernoulli distribution with success probability $p$ and $\expit(x) = 1/(1 + \exp(-x))$. The observation probability is:
	      \[
		      \mathbb{P}(W_{ij} = 1 \mid X_{ij}) =  \frac{\expit(2 X_{ij} - 1)}{1 + 0.1 \exp(Y_{ij})},
	      \]
	      where $Y_{ij} $ is generated from standard normal distribution independently.

	\item \textbf{DGP2:} $ X_{ij} = m_{ij} + \mathcal{N}(0,1) $, with observation probability:
	      \[
		      \mathbb{P}(W_{ij} = 1 \mid X_{ij}) =  \frac{\expit(2 X_{ij})}{1 + 0.1  \exp(Y_{ij})}.
	      \]
\end{list}

We take \(n\) as $50, 100, 200 $ and $400 $ to compare the computing time and estimation accuracy across different methods. Note that with this type of data generation, we have \(\bm{M} \approx \mathcal{T}(\bm{M})\), as the row and column spaces of \(\bm{M}\) are almost orthogonal to \(\bm{1}_{n}\). Therefore, from Theorem \ref{th:thransform}, we can infer that our estimator \(\hat{\bm{M}}\) \eqref{eq:nuclearpenalty} is close to the true value \(\bm{M}\).

Here we denote our method as RCU (Row- and Column-wise matrix U-statistic) method, with the FISTA accelerated one as RCU$_{\mathrm{acce}} $, and compare our method with several other baseline methods: EMU (Entire Matrix U-statistic) \cite{li2024a}, MHT \cite{MHT10}, NW \cite{NW12}, MAX \cite{MNC16}, MWC \cite{MCL21}, SBJ1 \cite{ILM20}, and SBJ2, which modifies SBJ1 by utilizing matrices \(\bm{X}^\top\) and \(\bm{W}^\top\). Here we exclude the case $n = 400$ for EMU as it's too time-consuming.  For each method, we perform 100 iterations repeat the simulation 50 times, report the computing time and the mean and standard deviation of the RMSE (Root Mean Square Error). For brevity, we present only the results for $n = 200$ and $400$ in the tables. See the Supplement Material Section S.5.1 for detailed descriptions of all baseline methods.

\textbf{DGP1:}

\begin{figure}[h]
	\centering
	\begin{subfigure}[b]{0.3\textwidth}
		\includegraphics[width=\textwidth]{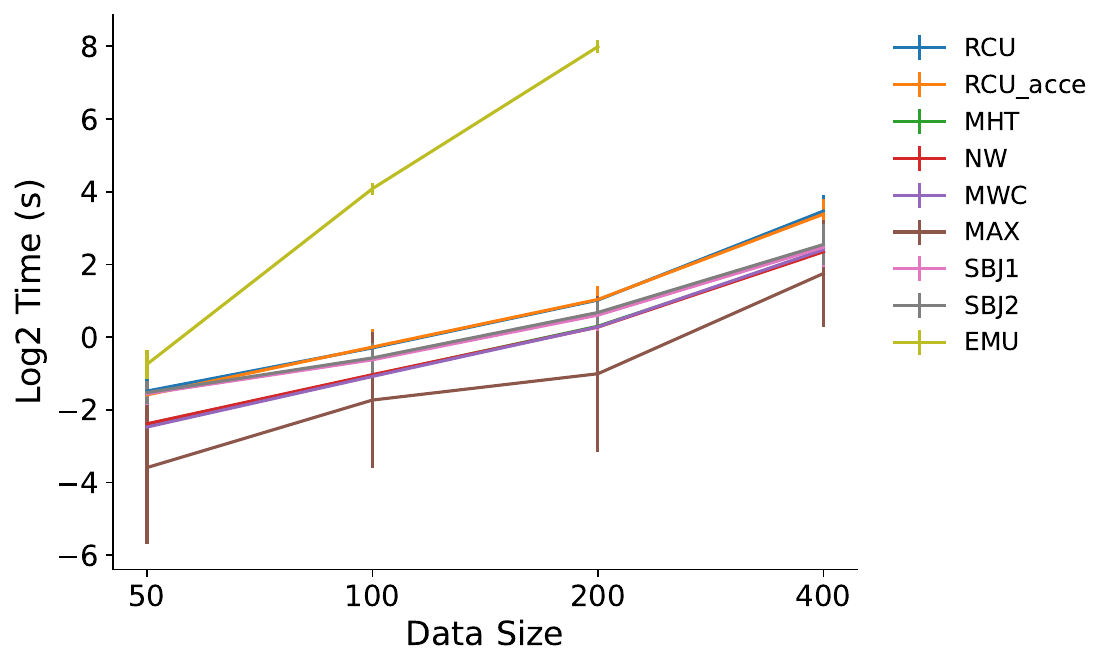}
		\caption{Log2-transformed computation time-varying sample sizes}
		\label{fig:time_log}
	\end{subfigure}
	\hfill
	\begin{subfigure}[b]{0.3\textwidth}
		\includegraphics[width=\textwidth]{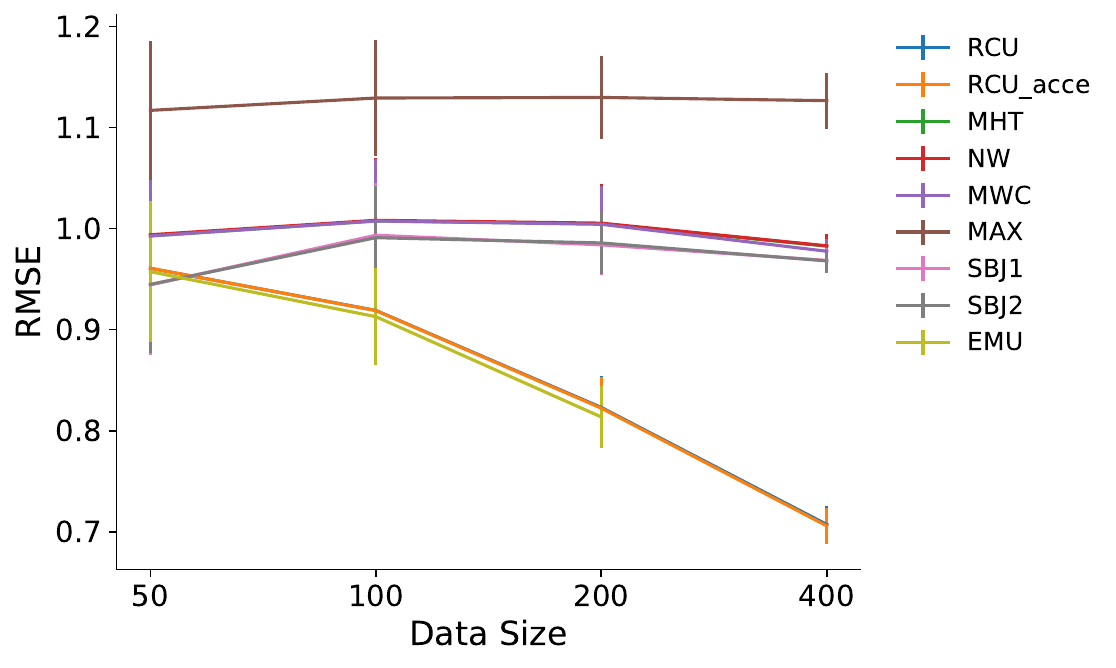}
		\caption{RMSE values for different methods across varying sample sizes}
		\label{fig:rmse_log}
	\end{subfigure}
	\hfill
	\begin{subfigure}[b]{0.3\textwidth}
		\includegraphics[width=\textwidth]{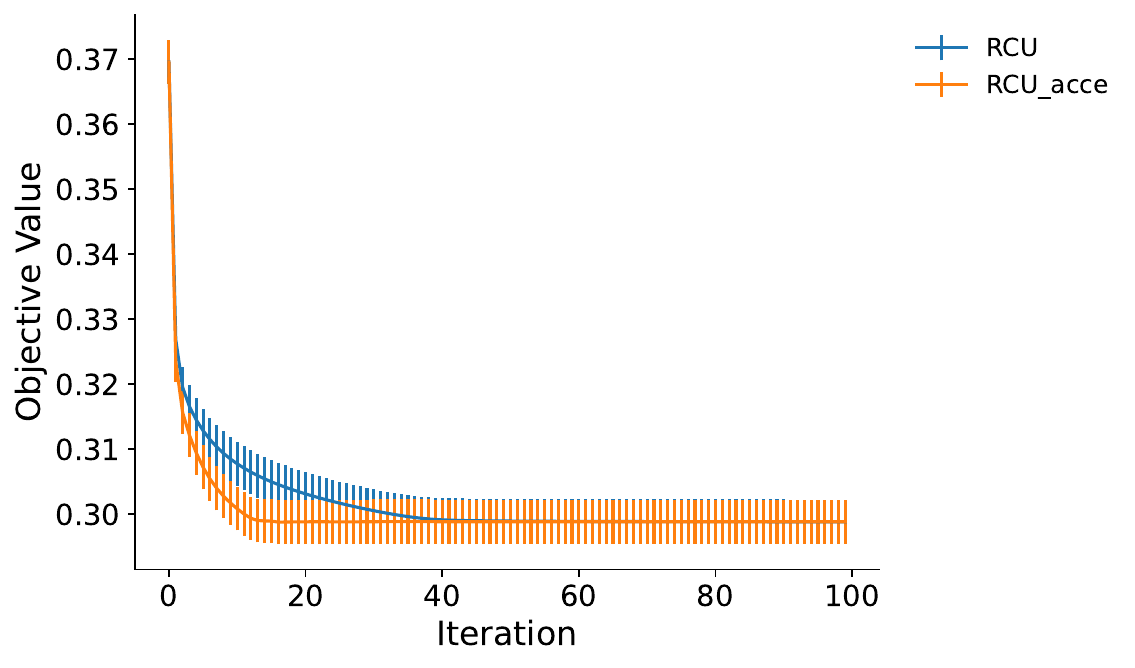}
		\caption{Objective value varying the number of iterations when $n = 400$}
		\label{fig:acce_log}
	\end{subfigure}
	\caption{Errorbar plots for DGP1}
	\label{fig:threefigures1}
\end{figure}

As shown in Table \ref{tab:dgp1}, the RMSE of our method is comparable to that of EMU (around $1\% $ worse than EMU), but the computing time is significantly reduced. The RMSE of RCU$_{\mathrm{acce}}$ is marginally lower than that of RCU, as it incorporates computational acceleration. Figure \ref{fig:acce_log} demonstrates that the RCU$_{\mathrm{acce}}$ algorithm consistently converges within approximately \(15\) iterations, indicating that FISTA variant achieves faster convergence.

Since the other methods do not account for the flexible nonignorable missing mechanism, their RMSE values are significantly higher, approximately \(1\). This is close to the RMSE obtained when using \(\bm{0}\) as the estimator, given that the variance of the elements in \(\bm{M}\) is \(1\).

As illustrated in Figure \ref{fig:time_log}, the computing time aligns with the computational complexity: EMU's time complexity is \(O(n^4)\), while the other methods are around \(O(n^3)\). Consequently, the slope of the logarithm of computation time with respect to sample size for EMU is much larger than that of the other methods. For instance, when \(n = 200\), EMU requires approximately \(250\) seconds, while RCU and RCU$_{\mathrm{acce}}$ only require around \(2.1\) seconds. Due to the rapid growth rate of EMU's computing time with sample size, it becomes impractical for high-dimensional matrix data. From our simulations, our method consistently requires approximately \(2\) times the runtime of MHT, the fastest baseline method, making it suitable for real-world high-dimensional matrix data applications.

\begin{table}[h]
	\caption{Mean and standard errors of RMSE and time spend for DGP1}
	\label{tab:dgp1}
	\centering
	\begin{adjustbox}{width = 1\textwidth}
		\begin{tabular}{l c c c c}
			\hline
			                    & \multicolumn{2}{c}{$n = 200 $} & \multicolumn{2}{c}{$n = 400 $}                                          \\
			Method              & RMSE                           & Time Spend                     & RMSE              & Time Spend         \\
			\hline
			RCU                 & 0.8230$\pm$0.0309              & 2.1015$\pm$0.5934              & 0.7072$\pm$0.0183 & 11.5693$\pm$3.5451 \\
			RCU$_\mathrm{acce}$ & 0.8223$\pm$0.0308              & 2.1192$\pm$0.5538              & 0.7058$\pm$0.0181 & 10.8758$\pm$3.2899 \\
			EMU                 & 0.8136$\pm$0.0304              & 256.2183$\pm$34.0583           &                   &                    \\
			MHT                 & 1.0047$\pm$0.0385              & 1.2786$\pm$0.3764              & 0.9826$\pm$0.0117 & 5.4838$\pm$2.0561  \\
			NW                  & 1.0053$\pm$0.0390              & 1.2593$\pm$0.3838              & 0.9829$\pm$0.0116 & 5.4540$\pm$2.1950  \\
			MWC                 & 1.0042$\pm$0.0380              & 1.2515$\pm$0.3112              & 0.9775$\pm$0.0118 & 5.7714$\pm$2.4983  \\
			MAX                 & 1.1297$\pm$0.0407              & 2.3169$\pm$4.5252              & 1.1265$\pm$0.0276 & 7.6071$\pm$13.4083 \\
			SBJ1                & 0.9838$\pm$0.0295              & 1.5896$\pm$0.4886              & 0.9685$\pm$0.0118 & 5.9679$\pm$2.3524  \\
			SBJ2                & 0.9858$\pm$0.0307              & 1.6660$\pm$0.4980              & 0.9680$\pm$0.0115 & 6.3163$\pm$2.5580  \\
			\hline
		\end{tabular}
	\end{adjustbox}
\end{table}

\textbf{DGP2:}

\begin{figure*}
	\centering
	\begin{subfigure}[b]{0.3\textwidth}
		\includegraphics[width=\textwidth]{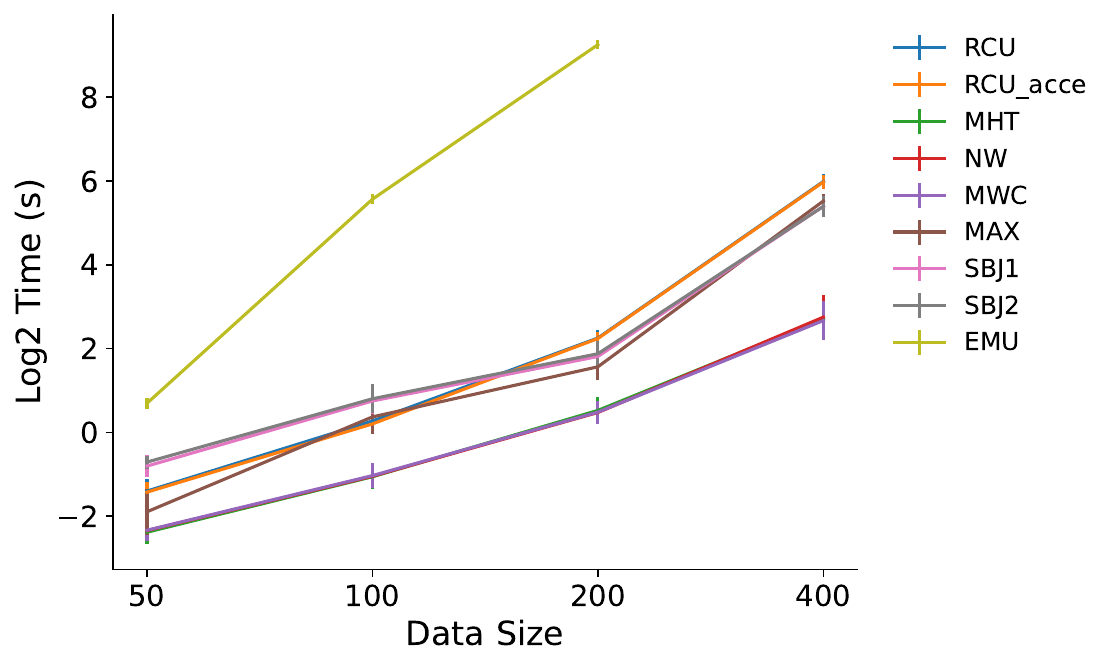}
		\caption{Log2-transformed computation time-varying sample sizes}
		\label{fig:time_gau}
	\end{subfigure}
	\hfill
	\begin{subfigure}[b]{0.3\textwidth}
		\includegraphics[width=\textwidth]{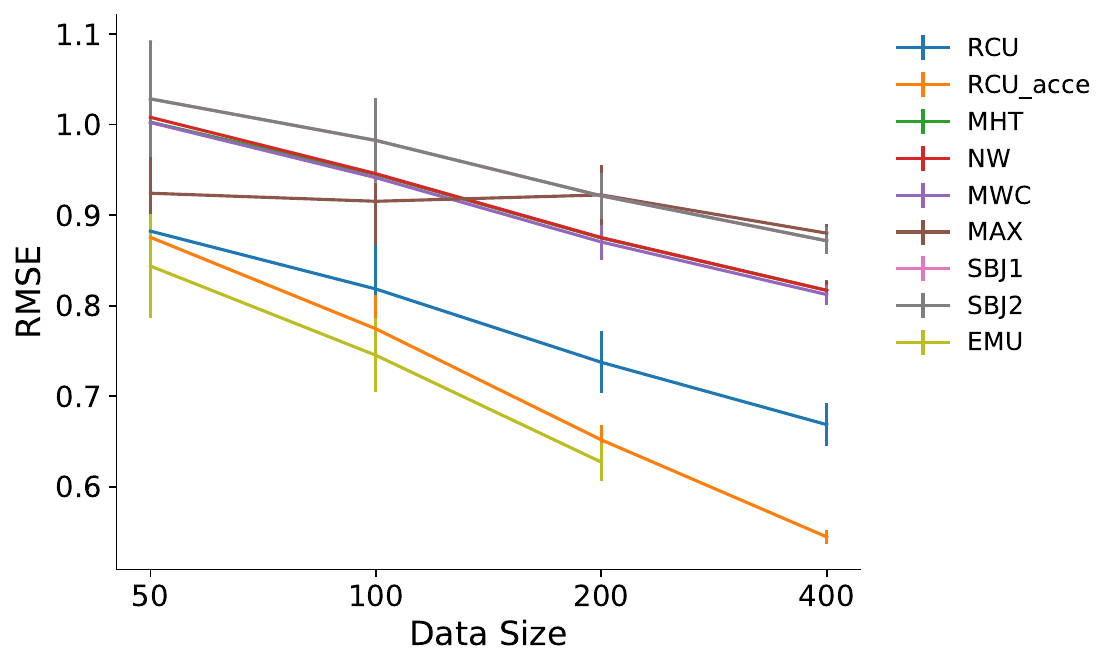}
		\caption{RMSE values for different methods across varying sample sizes}
		\label{fig:rmse_gau}
	\end{subfigure}
	\hfill
	\begin{subfigure}[b]{0.3\textwidth}
		\includegraphics[width=\textwidth]{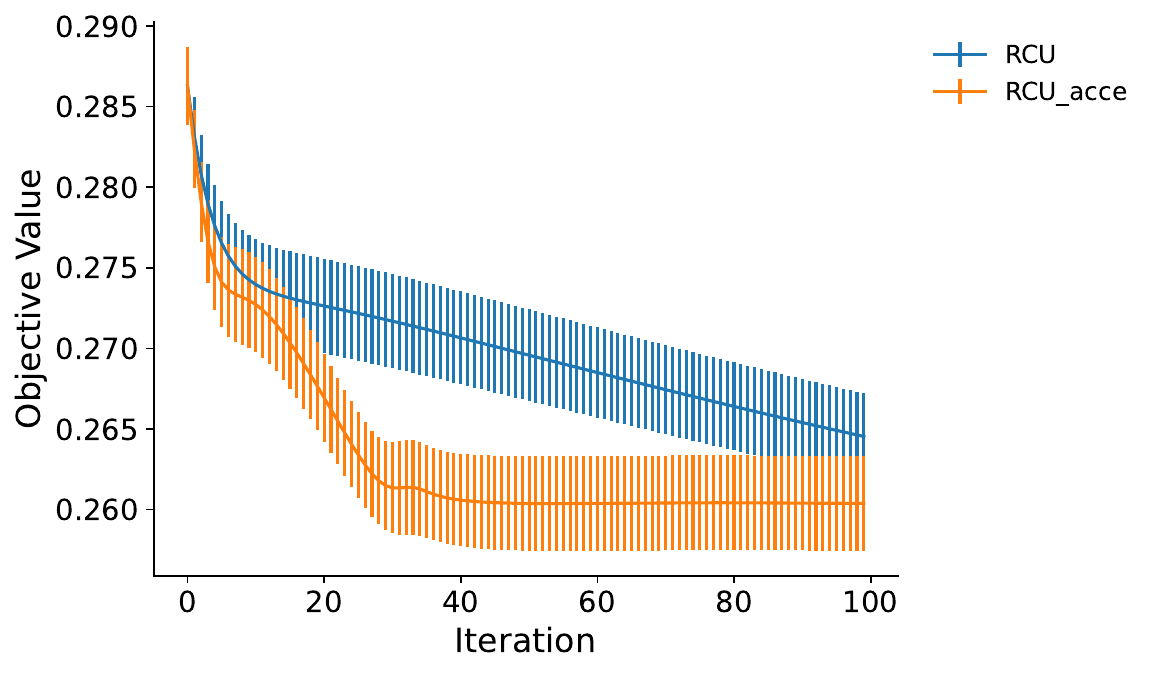}
		\caption{Objective value varying the number of iterations when $n = 400$}
		\label{fig:acce_gau}
	\end{subfigure}
	\caption{Errorbar plots for DGP2}
	\label{fig:threefigures2}
\end{figure*}

As shown in Table \ref{tab:dgp2}, the RMSE of RCU$_{\mathrm{acce}}$ is approximately \(3.9\%\) worse than that of EMU, but the computation time is significantly faster. Figure \ref{fig:time_gau} shows similar results to Figure \ref{fig:time_log}, where the slope of EMU is much steeper than those of the other methods. Additionally, Figure \ref{fig:rmse_gau} demonstrates that the comparison methods, which do not account for the flexible nonignorable missing mechanism, are unsuitable for DGP2. It also shows that the RMSE of RCU is consistently higher than that of RCU$_{\mathrm{acce}}$. The reason for this is illustrated in Figure \ref{fig:acce_gau}: RCU$_{\mathrm{acce}}$ converges within approximately \(40\) iterations, while RCU fails to converge within \(100\) iterations as: \(\mu = L_w L_x \approx 5.5\) being too large compared to \(1.1\) -- the value $\mu $ used for other baseline methods for DGP2 and all methods for DGP1.

Comparing Tables~\ref{tab:dgp1} and~\ref{tab:dgp2}, we observe significant increases in computation time for several methods. The SBJ1 and SBJ2 methods require 2--8 times longer to compute under DGP2 compared to DGP1, as the posterior distribution admits a closed-form solution for DGP1 but requires sampling-based approximation for DGP2. Similarly, RCU, RCU$_{\mathrm{acce}}$, and EMU exhibit 2--7 times longer computation times for DGP2. This difference arises because DGP1 contains numerous identical pairs ($X_{ij} = X_{i'j'}$) that avoid the gradient computations of $l_{ij,i'j'}$ (equals zero), while DGP2's continuous $X_{ij}$ values eliminate this computational shortcut. These results demonstrate our method's more suitable for binary matrix completion scenarios compared to continuous data.

\begin{table}[h]
	\caption{Mean and standard errors of RMSE and time spend for DGP2}
	\label{tab:dgp2}
	\centering
	\begin{adjustbox}{width = 1\textwidth}
		\begin{tabular}{l c c c c c c c c}
			\hline
			                    & \multicolumn{2}{c}{$n = 200 $} & \multicolumn{2}{c}{$n = 400 $}                                          \\
			Method              & RMSE                           & Time Spend                     & RMSE              & Time Spend         \\
			\hline
			RCU                 & 0.7375$\pm$0.0341              & 4.8060$\pm$0.6078              & 0.6686$\pm$0.0235 & 64.0133$\pm$7.4647 \\
			RCU$_\mathrm{acce}$ & 0.6518$\pm$0.0168              & 4.7594$\pm$0.5126              & 0.5447$\pm$0.0081 & 63.5143$\pm$7.2761 \\
			EMU                 & 0.6272$\pm$0.0212              & 613.5168$\pm$47.0800           &                   &                    \\
			MHT                 & 0.8748$\pm$0.0199              & 1.4736$\pm$0.3325              & 0.8169$\pm$0.0109 & 6.9613$\pm$2.3355  \\
			NW                  & 0.8752$\pm$0.0197              & 1.4149$\pm$0.2619              & 0.8169$\pm$0.0108 & 7.1953$\pm$2.6424  \\
			MWC                 & 0.8703$\pm$0.0202              & 1.4234$\pm$0.2612              & 0.8123$\pm$0.0110 & 6.7053$\pm$2.1629  \\
			MAX                 & 0.9223$\pm$0.0328              & 3.0641$\pm$1.1334              & 0.8801$\pm$0.0096 & 46.3565$\pm$5.5798 \\
			SBJ1                & 0.9214$\pm$0.0251              & 3.5692$\pm$0.6145              & 0.8717$\pm$0.0150 & 42.7706$\pm$7.1220 \\
			SBJ2                & 0.9213$\pm$0.0252              & 3.7531$\pm$0.7502              & 0.8716$\pm$0.0150 & 42.7568$\pm$7.3209 \\
			\hline
		\end{tabular}
	\end{adjustbox}
\end{table}

For comprehensive analysis of the sensitivity analysis to parameters $(\lambda, \alpha)$ and the performance of our estimator under varying missing rates compared with alternative methods, please see Supplement Material S.5.2.

\subsection{Real Data Analysis}

For recommendation systems, the accuracy of rating ranking estimation is crucial for determining which items to recommend to users. To assess ranking performance, we employ three quantitative metrics: \(\mathrm{RANK}_1\), \(\mathrm{RANK}_2\), and \(\mathrm{RANK}_3\) \cite{CFA08}:

\begin{itemize}
	\item \textbf{$\mathrm{RANK}_1$}: The row-wise expected percentile ranking proposed by \cite{CFA08}:
	      \[
		      \mathrm{RANK}_1 = \frac{\sum_{(i,j) \in \text{test set}} X_{ij} \times \mathrm{rank}_{1,ij}}{\sum_{(i,j) \in \text{test set}} X_{ij}},
	      \]
	      where $\mathrm{rank}_{1,ij}$ is the predicted percentile rank of item $j$ for user $i$ among all $\hat{m}_{ij}$, $1 \leq j \leq n_2$ (the same row), and $X_{ij}$ is the corresponding value in the test set.
	      For example, if the predicted value $\hat{m}_{ij}$ is the highest among all $\hat{m}_{ij}$, $1 \leq j \leq n_2$, then $\mathrm{rank}_{1,ij} = 0$; conversely, if $\hat{m}_{ij}$ is the lowest, then $\mathrm{rank}_{1,ij} = 1$.

	\item \textbf{$\mathrm{RANK}_2$}: The column-wise expected percentile ranking, which is a modification of the above metric. Here, $\mathrm{rank}_{2,ij}$ replaces $\mathrm{rank}_{1,ij}$, where $\mathrm{rank}_{2,ij}$ is the predicted percentile rank of item $j$ for user $i$ among all $\hat{m}_{ij}$, $1 \leq i \leq n_1$ (the same column).

	\item \textbf{$\mathrm{RANK}_3$}: The overall expected percentile ranking, where $\mathrm{rank}_{3,ij}$ -- the predicted percentile rank of user $i$ for item $j$ among all $\hat{m}_{ij}$, $1 \leq i \leq n_1$, $1 \leq j \leq n_2$ (over the full matrix) replaces $\mathrm{rank}_{1,ij}$.
\end{itemize}

As demonstrated by the calculation of these metrics, smaller value indicates better ranking estimation, and the expected value for a completely randomized matrix is \(50\%\).

In this section, we apply our proposed method to two real-world data sets: the Learning from Sets of Items data\footnote{The movie rating data can be downloaded from \url{https://grouplens.org/data sets/learning-from-sets-of-items-2019/}.} and the Senate Voting data\footnote{The detailed voting records are documented on the website \url{https://www.senate.gov/legislative/votes_new.htm}.}. We compare our method with the baseline methods discussed in the simulation section, excluding EMU, as it cannot handle high-dimensional matrix data. For each data set, we randomly split the data into training and test sets with an 80\%/20\% ratio and report the performance over 50 times repeat.

\subsubsection{Learning from Sets of Items Data}

We utilize the movie rating data set collected from \url{https://movielens.org} between February and April 2016 \cite{MLD19}, comprising 458,970 ratings on a scale of 0.5 to 5 by 854 users for 13,012 movies. For our analysis, we focus on the 1,000 most popular movies, which have received the highest number of ratings. This submatrix is $854 \times 1000 $ with 231,296 items. We define $X_{ij} = 1$ if the rating is no less than 4, indicating the user's preference for the movie, and $X_{ij} = 0$ otherwise.

\begin{figure}[h]
	\centering
	\begin{subfigure}[b]{0.3\textwidth}
		\includegraphics[width=\textwidth]{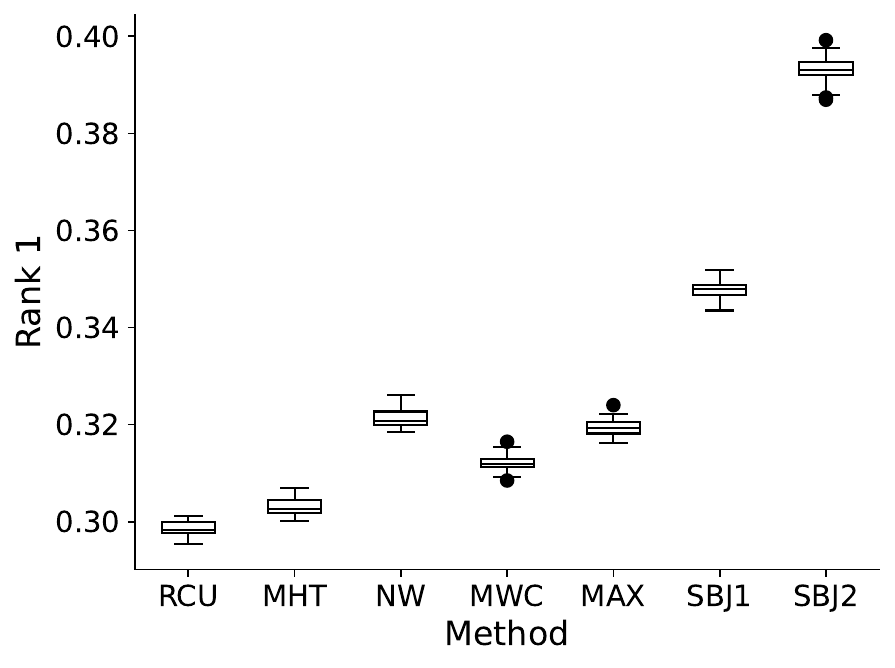}
	\end{subfigure}
	\hfill
	\begin{subfigure}[b]{0.3\textwidth}
		\includegraphics[width=\textwidth]{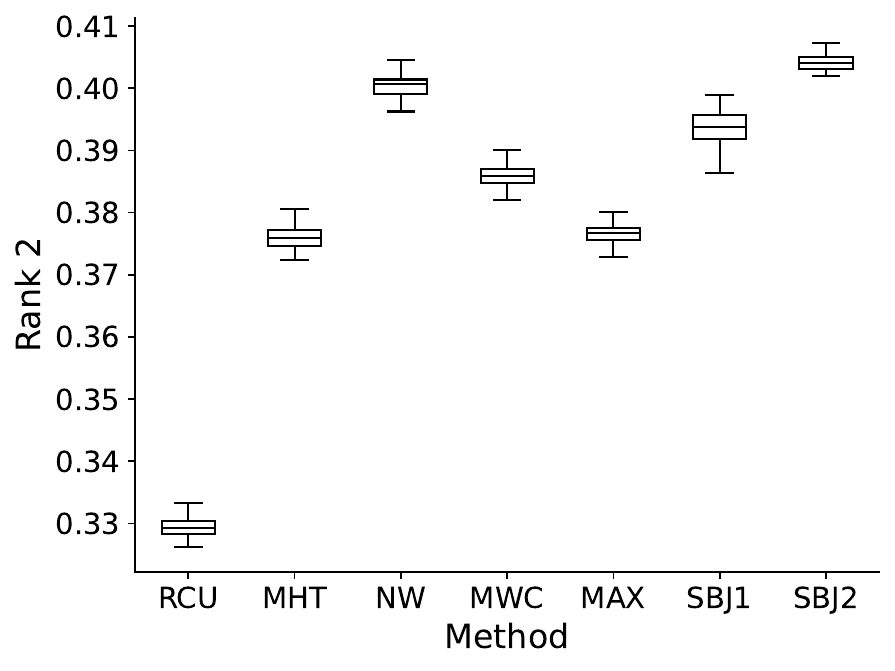}
	\end{subfigure}
	\hfill
	\begin{subfigure}[b]{0.3\textwidth}
		\includegraphics[width=\textwidth]{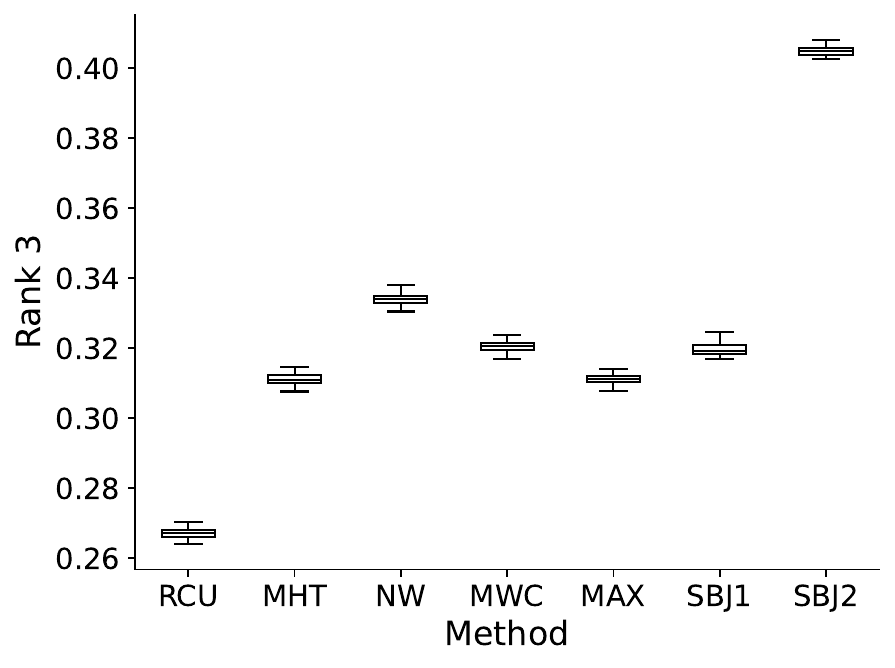}
	\end{subfigure}
	\caption{Box plot of ranking value for learning from sets of items data set, with plots for \(\mathrm{RANK}_1\), \(\mathrm{RANK}_2\), and \(\mathrm{RANK}_3\) from left to right correspondingly}
	\label{fig:data1}
\end{figure}

The numerical results are presented in Table \ref{tab:data1} and Figure \ref{fig:data1}. The results demonstrate that our method achieves optimal ranking performance across all metrics. For the row-wise ranking, our method performs slightly better than the MHT method, while for the column-wise and overall ranking, it significantly outperforms the other methods.

\begin{table}[h]
	\caption{Mean and standard errors of ranking value for learning from sets of items data set}
	\label{tab:data1}
	\centering
	\begin{tabular}{lccc}
		\hline
		     & Rank 1                     & Rank 2                     & Rank 3                     \\
		\hline
		RCU  & \textbf{0.2986}$\pm$0.0015 & \textbf{0.3293}$\pm$0.0017 & \textbf{0.2671}$\pm$0.0015 \\
		MHT  & 0.3030$\pm$0.0017          & 0.3759$\pm$0.0018          & 0.3110$\pm$0.0017          \\
		NW   & 0.3212$\pm$0.0017          & 0.4003$\pm$0.0017          & 0.3339$\pm$0.0017          \\
		MWC  & 0.3121$\pm$0.0017          & 0.3859$\pm$0.0018          & 0.3205$\pm$0.0017          \\
		MAX  & 0.3194$\pm$0.0015          & 0.3766$\pm$0.0016          & 0.3111$\pm$0.0015          \\
		SBJ1 & 0.3479$\pm$0.0018          & 0.3937$\pm$0.0029          & 0.3197$\pm$0.0018          \\
		SBJ2 & 0.3930$\pm$0.0027          & 0.4041$\pm$0.0013          & 0.4049$\pm$0.0013          \\
		\hline
	\end{tabular}
\end{table}

\subsubsection{Senate Voting Data}
We apply our proposed method to United States Senate roll call voting data spanning the 111th--113th Congresses (January 11, 2009--December 16, 2014). Following the data preprocessing described in \cite{SIB23}, we analyze 159,186 votes across 1,648 bills from 139 senators. We define $X_{ij} = 1$ when senator $i$'s vote on bill $j$ aligns with the Republican Party position, and $X_{ij} = 0$ otherwise. The value of $X_{ij}$ is considered missing if the senator chose not to vote or was absent.

\begin{figure}[h]
	\centering
	\begin{subfigure}[b]{0.3\textwidth}
		\includegraphics[width=\textwidth]{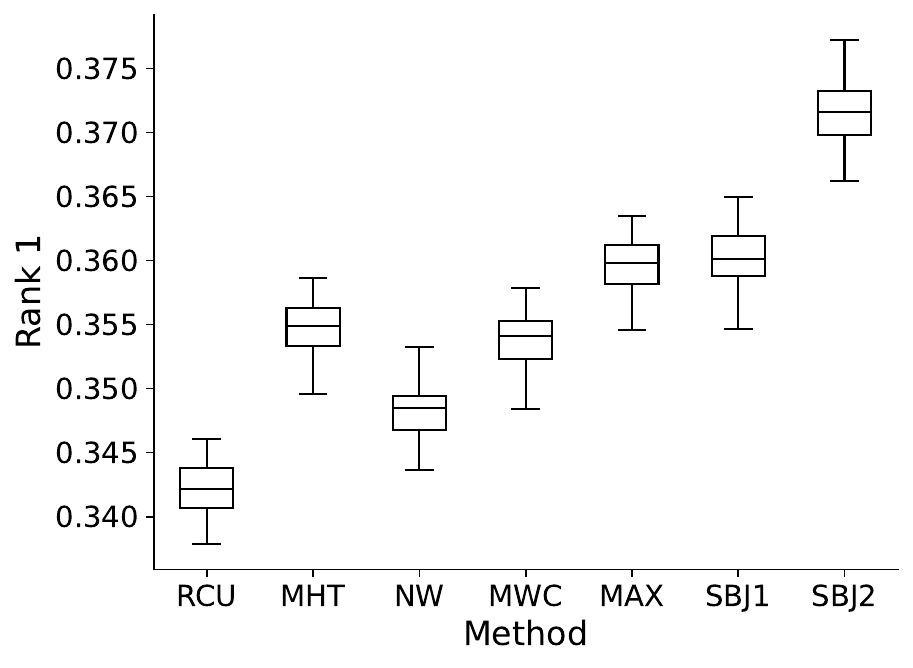}
	\end{subfigure}
	\hfill
	\begin{subfigure}[b]{0.3\textwidth}
		\includegraphics[width=\textwidth]{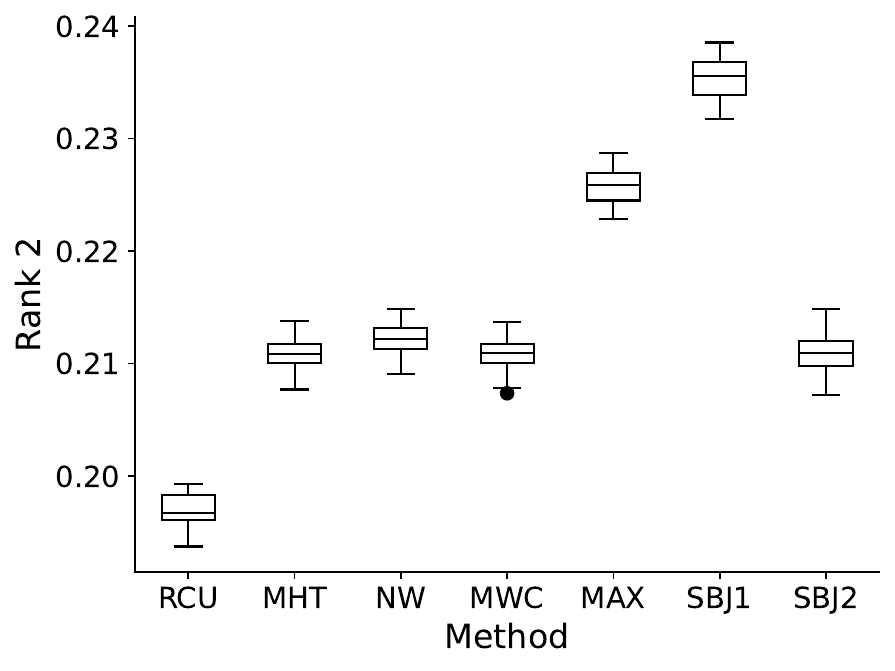}
	\end{subfigure}
	\hfill
	\begin{subfigure}[b]{0.3\textwidth}
		\includegraphics[width=\textwidth]{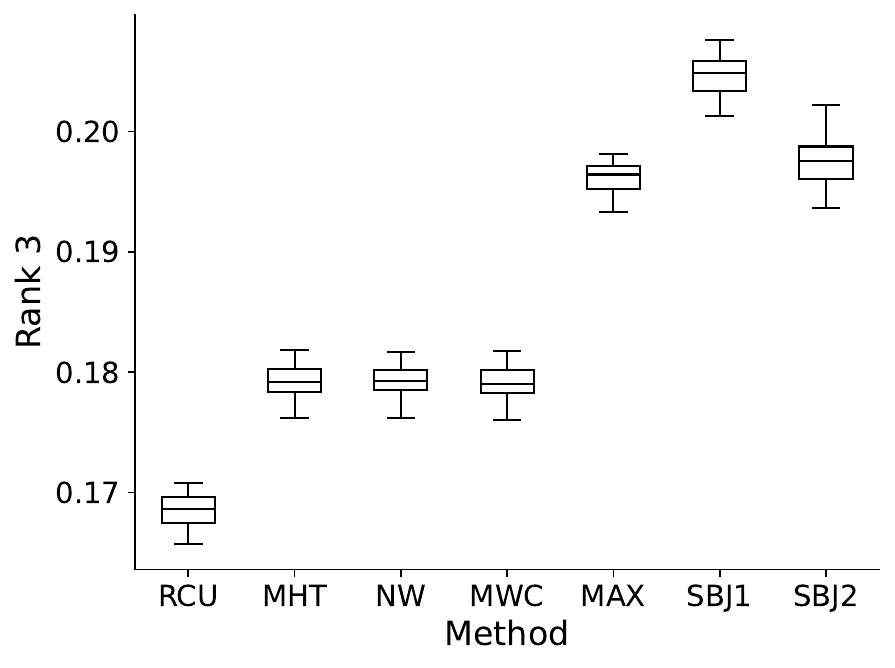}
	\end{subfigure}
	\caption{Box plot of ranking value for Senate Vote data set.}
	\label{fig:data3}
\end{figure}

The numerical results are presented in Table~\ref{tab:data3} and Figure~\ref{fig:data3}. Our method consistently and significantly outperforms the other methods across all three metrics.
\begin{table}[h]
	\caption{Ranking value result for Senate Vote data set}
	\label{tab:data3}
	\centering
	\begin{tabular}{lccc}
		\hline
		     & Rank 1                     & Rank 2                     & Rank 3                     \\
		\hline
		RCU  & \textbf{0.3421}$\pm$0.0020 & \textbf{0.1969}$\pm$0.0014 & \textbf{0.1685}$\pm$0.0012 \\
		MHT  & 0.3547$\pm$0.0021          & 0.2108$\pm$0.0015          & 0.1792$\pm$0.0012          \\
		NW   & 0.3482$\pm$0.0020          & 0.2121$\pm$0.0014          & 0.1792$\pm$0.0012          \\
		MWC  & 0.3538$\pm$0.0020          & 0.2107$\pm$0.0015          & 0.1791$\pm$0.0012          \\
		MAX  & 0.3596$\pm$0.0022          & 0.2257$\pm$0.0015          & 0.1962$\pm$0.0012          \\
		SBJ1 & 0.3604$\pm$0.0023          & 0.2353$\pm$0.0017          & 0.2047$\pm$0.0015          \\
		SBJ2 & 0.3717$\pm$0.0025          & 0.2109$\pm$0.0016          & 0.1975$\pm$0.0019          \\
		\hline
	\end{tabular}
\end{table}

As demonstrated in the analyses of these data sets, incorporating the nonignorable missing mechanism consistently results in better ranking performance. This indicates that our proposed method is more robust against complex real-world missing mechanisms.

\section{Conclusion}
\label{sec:conclusion}

In this paper, we propose a computationally efficient estimator for high-dimensional matrix completion under generalized nonignorable missing mechanisms. Our theoretical analysis and numerical experiments demonstrate that: (i) the estimator maintains computational efficiency comparable to classical MCAR/MAR matrix completion methods; (ii) it achieves similar theoretical error bounds while accommodating the more general MNAR mechanism; and (iii) it provides theoretically-grounded solutions to the generalized nonignorable missingness problem, with comprehensive simulations and real-data applications confirming its effectiveness in handling complex missing-data mechanisms.

While the proposed method advances the current methodology, several important extensions remain. Future research directions include: (i) relaxing structural assumptions on the missing mechanism \eqref{eq:missingmechanism}; (ii) developing statistical inference procedures following~\cite{xia2021statistical}; and (iii) further reducing computational complexity through U-statistic reduction~\cite{shaoUstatisticReductionHigherorder2025} and truncated matrix SVD techniques~\cite{SVD11}. These challenging yet important directions merit further investigation.

We expect this work to stimulate additional research on nonignorable missing data in high-dimensional matrix completion and to provide valuable insights for related statistical learning problems.

\bigskip
\begin{center}
	{\large\bf Supplement Material}
\end{center}

\begin{description}

	\item[Matrix\_U-statistic\_supp:] This Supplement Material contains: (i) extended discussion of related literature, (ii) additional simulation studies, and (iii) complete proofs of all theoretical results presented in Sections~\ref{sec:modelandmethod} and~\ref{sec:theory} (PDF format).

	\item[Code, Data and Output:] Complete code for reproducing all simulation studies and real-data applications, along with corresponding output files and raw datasets.

\end{description}

\bibliographystyle{agsm}
\bibliography{reference.bib}
\end{document}